\pdfoutput=1
\documentclass{article}
\usepackage[T1]{fontenc}
\usepackage{iclr2027_conference,times}

\usepackage{amsmath,amsfonts,bm}

\def\eqref#1{equation~\ref{#1}}

\def\1{\bm{1}}

\DeclareMathAlphabet{\mathsfit}{\encodingdefault}{\sfdefault}{m}{sl}
\SetMathAlphabet{\mathsfit}{bold}{\encodingdefault}{\sfdefault}{bx}{n}

\usepackage{hyperref}
\usepackage{url}
\usepackage{booktabs}
\usepackage{multirow}
\usepackage{array}
\usepackage{graphicx}
\usepackage{amsmath}
\usepackage{xcolor}
\usepackage{colortbl}
\usepackage{enumitem}
\usepackage{wrapfig}
\usepackage{needspace}
\usepackage{tikz}
\usetikzlibrary{arrows.meta,backgrounds}

\newcommand{\xtree}{X-Tree}
\newcommand{\xscore}{$\xscoresym$-Score}
\newcommand{\xscoresym}{\mathcal{X}}
\newcommand{\ctok}[2]{\texttt{#1}$\langle$\texttt{#2}$\rangle$}
\newcommand{\ctokx}[2]{\texttt{#1}$\langle$#2$\rangle$}

\title{X-Tree: Tokenizing Reusable Experience for Efficient Agent Generalization}

\author{Sitao Cheng$^{1}$, Xunjian Yin$^{2}$, Zhiyuan Sun$^{1}$, Yuxuan Li$^{1}$, \\
\textbf{Ruiwen Zhou$^{3}$, Xiangru Jian$^{1}$ \& Victor Zhong$^{1}$} \\
$^{1}$University of Waterloo \quad $^{2}$Duke University \quad $^{3}$National University of Singapore \\
\texttt{\{sitao.cheng,victor.zhong\}@uwaterloo.ca} \\[2pt]
Project page: \url{https://sitaocheng.github.io/xtree} \\
Code: \url{https://github.com/sitaocheng/X-Tree}
}

\iclrfinalcopy

\begin{document}

\maketitle
\lhead{Preprint}

\begin{abstract}
Multi-step agents are trained on flat action streams: SFT and RLVR weight every token uniformly and ignore the sub-procedures that recur across tasks, the hierarchy that lets humans plan top-down from reusable routines.
This structure sits unused, and flat training uses each scarce trajectory less fully than its content allows.
Recent agents do use that structure, but only as LLM-written skills in context, never in the weights, so their gains do not generalize beyond retrieval.
We instead recover this hierarchy from the data itself and train on it, with no LLM calls.
Following text tokenizers, which build a vocabulary by counting alone, we score action spans by reusability and merge canonicalized actions into a \emph{reusable eXperience tree} (\xtree{}). Each \xtree{} node captures how a frequent and success-bearing skill is composed from sub-skills, guiding efficient generalization.
We integrate \xtree{} into three training settings: offline RL, with each node as a training instance; online RLVR, with an adaptive skill bonus; and on-policy self-distillation, with \xtree{} as the self-teacher's privileged context.
Across WebArena, ScienceWorld, and WebShop at three model scales, \xtree{} improves over standard recipes at matched data and budget by 
up to $4.5\%$ SR on WebArena, $5.8\%$ SR on ScienceWorld and $4.1\%$ success on WebShop.
Matched analyses attribute the gains to the \xtree{} structure and the three integrations.
\end{abstract}

\section{Introduction}
\label{sec:intro}

Human learners spontaneously discover hierarchies that compose recurring action sequences,
acting from goals through sub-goals to primitives
\citep{miller1956magical,botvinick2009hierarchically}.
Language models trained to operate websites \citep{zhou2023webarena}, resolve repository issues \citep{jimenez2024swebench}, and act over long-horizon interactive environments \citep{NEURIPS2024_5d413e48,xi2024agentgym} learn the opposite way.
A trajectory enters training, whether SFT or RLVR, as a flat stream of actions, and the objective weights every token equally \citep{ouyang2022training,shao2024deepseekmath}.
This equal weighting overlooks the sub-procedures that recur across tasks, such as browse-and-select or pick-up-and-move.
The hierarchy is present in the data, but the training objective does not use it.
Uniform training therefore extracts less from each trajectory than the data allows, requiring more data to learn.

This underutilization is especially costly because experience is scarce.
A trajectory is collected by a human \citep{wang2026opencua} or synthesized and verified by a model \citep{gandhi2025gobrowse}.
An environment must be built, its tasks written, and its verifier made trustworthy.
None of this scales the way web text does.
We therefore ask \textbf{\textit{how to extract the most from a finite set of experience}}.

To extract more insight from a given pool of experience, recent agent harnesses such as OpenClaw ask an LLM to summarize past experience as natural-language workflows and replay them \emph{in context} \citep{wang2023voyager,wang2024awm}.
Nothing is compressed into the weights, so the experience cannot generalize beyond retrieval.
On-policy self-distillation (OPSD) instead brings such skills into training, where a self-teacher reads them as privileged context and re-weights the learner's own tokens \citep{zhao2026self,lu2026sdar}.
Both depend on an LLM to write the abstraction, which is costly, uncontrollable and leaves a library which cannot be reproduced from the data alone.

\begin{figure}
    \centering
    \includegraphics[width=\linewidth]{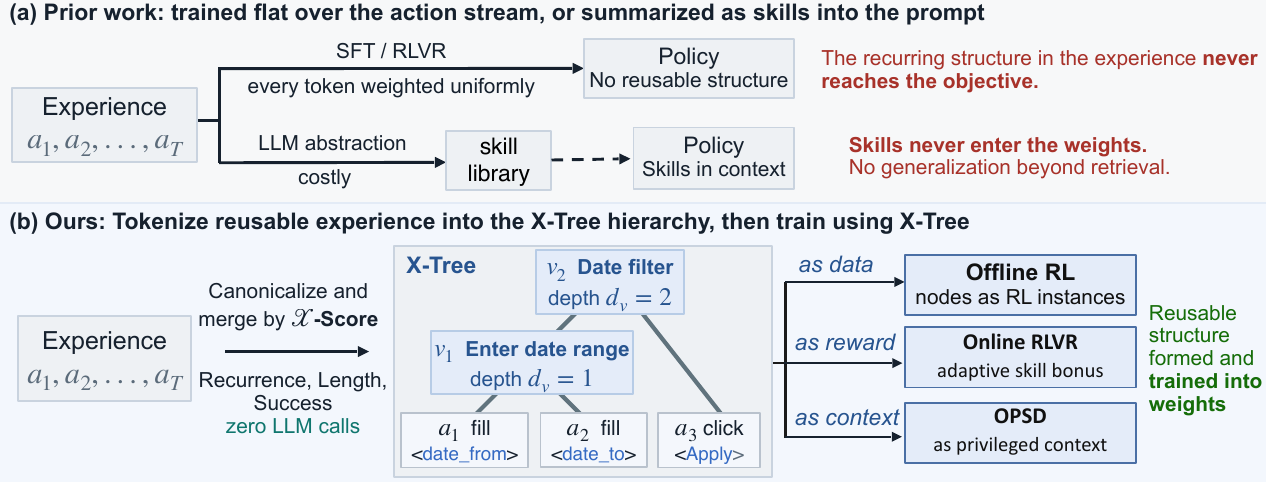}
    \caption{\textbf{Training with experience.} (a) Prior training weights
every token uniformly. Harness systems use LLM-written skills in context, never in the weights.
(b) We mine reusable experience by \xscore{} (measuring recurrence, length, success) into the \xtree{} hierarchy with zero LLM calls. Each node $v$ is a recurring, success-bearing structure with depth $d_v$.
We train \xtree{} into model weights by integrating \xtree{} as data (offline RL), as reward
(online RLVR) and as context (OPSD). Figure \ref{fig:trajectory-tree} demonstrates a concrete \xtree{} example from a trajectory.}
    \label{fig:concept}
\end{figure}

Language models already solve the analogous problem for text before learning begins.
A recurrence-based tokenizer merges co-occurring characters into a structured vocabulary \citep{sennrich2016bpe}.
In this paper, \textit{we tokenize experience into reusable structures} (Figure~\ref{fig:concept}).
We canonicalize each action step in a trajectory into a typed token (\textit{i.e.,} verb$<$role$>$).
We introduce \xscore{}, a measure of reusability that scores a span by how often it recurs, how long it is, and how often it appears in successful episodes.
Guided by \xscore{}, we merge adjacent pairs into \textit{nodes}, then merge nodes into larger nodes hierarchically.
This produces \textbf{reusable eXperience tree} (\textbf{\xtree{}}), a deterministic, auditable hierarchy over a trajectory corpus built with zero LLM calls. Each \xtree{} node captures \textit{how a frequent and success-bearing skill is composed from sub-skills,} guiding efficient generalization.

Our preliminary result shows that \xtree{} is already useful as in-context advice (Appendix~\ref{app:prompted}).
We integrate \xtree{} into three training settings (Figure~\ref{fig:exemplar}).
\textbf{1) Offline RL,} when only trajectories exist without an environment.
Each \xtree{} node acts as one training instance. The policy partially rolls out from the gold prefix that precedes the node.
We optimize with GRPO over a step-matching reward with a depth scaled completion bonus.
\textbf{2) Online RLVR,} when an environment and a verifier exist.
\xtree{} adds a bonus for each \xtree{} skill a rollout executes with an adaptive weight that supplies signal while verifier signal is scarce and down-weights it as the signal becomes informative.
\textbf{3) OPSD,} where \xtree{} replaces the LLM-written skill bank that the self-teacher reads.

We experiment across three environments and three model scales at matched data and budget.
Our offline RL with \xtree{} gains $4.5\%$ success rate (SR) over full-data SFT on WebArena \citep{zhou2023webarena} with Qwen2.5-7B. Ablations, such as random tree, attribute the gain to \xtree{} and our RL method.
Our online RLVR with \xtree{} stays ahead of outcome-based RL across all scales by up to $4.9\%$ SR, leading the held-out tasks, on ScienceWorld \citep{wang2022scienceworld}. It also consistently outperforms standard recipe by up to $3.6\%$ success and $4.6\%$ graded score on WebShop \citep{yao2022webshop}. For OPSD, \xtree{} as the privileged context of self-teacher matches LLM-written skills, with no LLM calls on both ScienceWorld and WebShop at all scales. 
We summarize our contributions:
\begin{itemize}[leftmargin=1.em, topsep=-1pt, itemsep=0pt]
\item \textbf{\xtree{}}, which tokenizes reusable experience from a trajectory pool into a deterministic hierarchy of skills with zero LLM calls. Each node captures the compositional signals to guide training.
\item Three integrations of \xtree{} with training, as data, reward, and context: offline RL for the trajectories-only setting; online RLVR with an adaptive, verifier-aligned skill bonus; and OPSD with \xtree{} as the self-teacher's privileged context (\S\ref{sec:method-offline}--\ref{sec:method-opsd}).
\item Experiments across three environments and three scales showing that \xtree{} improves over standard recipes at matched data and budget by a large margin, with ablations that attribute gains to \xtree{} structure and to the three integrations
(\S\ref{sec:results}).
\end{itemize}
\section{Related work}
\label{sec:related}

\textbf{Hierarchy learning.}
Hierarchical organization of
behavior is an established finding in cognitive science: capacity limits force recoding
into reusable structures, and learners act from goals through sub-goals to
primitives \citep{miller1956magical,gobet2001chunking,botvinick2009hierarchically}, choosing decompositions that trade utility against planning cost \citep{solway2014optimal,correa2023humans}. A long line of
work recovers such structure from demonstrations by segmenting trajectories into
options or sub-policies \citep{sutton1999options,konidaris2012cst,fox2017ddo,shiarlis2018taco,kipf2019compile,jiang2022love}.
We take from it the finding that bottom-up training on flat action streams
overlooks structure a learner should exploit. We recover the hierarchy as a
tree structure over actions handed to training rather than as a set of advice in context.

\textbf{LLM-induced skills and experience reuse.} 
Recent agents ask a
proprietary model to summarize reusable experience as natural-language workflows
\citep{wang2023voyager,wang2024awm,tang2025chemagent} or executable tools
\citep{zheng2025skillweaver,prabhu2026walt}, or store raw experience as verbal
reflections, distilled insights and retrieved trajectories
\citep{shinn2023reflexion,zhao2024expel,zheng2024synapse,su2025learnbyinteract}.
Both families keep the skills in prompt or a tool registry, costing LLM calls for abstraction. Skills are never compressed into the model, not generalizing beyond retrieval. We replace
the abstraction step with a deterministic miner and train the skills into the
weights.

\textbf{Compression as abstraction.} 
Grammar induction over symbol sequences
\citep{nevillmanning1997sequitur} and dictionary compression
\citep{larsson2000repair} are the classical ancestors. Library learning compresses programs under a description-length objective \citep{ellis2021dreamcoder,bowers2023stitch,grand2024lilo}. And robot policies tokenize continuous actions by vector quantization \citep{lee2024vqbet}. Closest to us, PRISE \citep{zheng2024prise} and Subwords-as-Skills
\citep{yunis2023subwords} apply Byte-Pair-Encoding (BPE) to quantized continuous-control actions. We
differ in alphabet (typed actions from a canonicalizer), in merge
criterion (\xscore{} to measure reusability, as behavior carries more information than text),
and in the use of vocabulary: not a fixed macro-action space but reusable experience as a prior
to guide training as data, as credit and as context.

\textbf{Multi-turn agent training.} Many domains lack
 an online environment and a task verifier \citep{zhou2023webarena,NEURIPS2024_5d413e48}. Our offline RL integration with \xtree{} provides a solution to train from the trajectory data \citet{gandhi2025gobrowse} alone. 
Recent studies train agents online with  multi-turn RL \citep{qi2024webrl,wei2025webagentr1,zhang2025rlvmr} using policy-gradient \citep{feng2025gigpo}.
Unlike a learned process-reward model, our shaping signal is mined from the trajectory pool itself. On the other hand, on-policy self-distillation trains a model toward its own predictions under a privileged context \citep{zhao2026self,shenfeld2026self,lu2026sdar}. We keep that objective and replace the LLM-written privileged context with the \xtree{}.

\section{\xtree{}: Tokenizing Reusable EXperience}
\label{sec:method}

Let a trajectory be a sequence of actions $\tau = a_1 a_2 \cdots a_T$
with an episode-level success label, and $\mathcal{D}$ be a pool of trajectories for a target domain. We build \xtree{} from $\mathcal{D}$
(\S\ref{sec:method-tokenizer}) and integrate it with training in three settings (Figure \ref{fig:exemplar}), \textit{i.e.,} Offline RL (\S\ref{sec:method-offline}), Online RLVR (\S\ref{sec:method-online}) and OPSD (\S\ref{sec:method-opsd}). We also integrate \xtree{} with SFT training in Appendix \ref{app:sft}.

\subsection{\xtree{} Mining}
\label{sec:method-tokenizer}

The text BPE tokenizer \citep{sennrich2016bpe} turns a character stream into a
vocabulary by recursively merging the most frequent adjacent pair, which is itself a hierarchy. We apply the idea to experience with two changes:
actions are first made comparable by canonicalization to reduce noise, and merges are chosen by how \emph{reusable}
the resulting pair is.

\textbf{Canonicalization.}
 We map each raw action (excluding thought) to a typed token \ctok{verb}{role}, where the role is a class or object the verb acts on, \textit{e.g.,} \texttt{fill(bid,'2/2/26')}$\mapsto$\ctok{type}{date}  (templates are in Appendix~\ref{app:canon}).
Other values in the action (\textit{e.g.,} element id, object name) are stripped.
Therefore, structurally identical
actions share a symbol, which makes recurrence visible.

\textbf{\xscore{}: measuring reusability of a merged skill.}
Recurrence-based text tokenization treats each occurrence of a character pair equally. Experience carries more signal: a pair of actions is reusable when it \emph{recurs} across the corpus, when it is \emph{long} enough
that reusing it induces better compression, and when it frequently appears in \emph{successful} episodes.

Let $u$ and $v$ be two symbols that occur adjacently in the corpus, each either
a canonicalized action or a node from an earlier merge, let $uv$ be the candidate node by merging them. Let $f_{uv}$ be the number of times the pair occurs adjacently, and $\ell_u$ be the number of primitive actions symbol $u$ expands to, so $\ell_u{+}\ell_v$ is the length of the candidate: a primitive action has $\ell{=}1$, merging \ctok{type}{text\_input} with
\ctok{type}{text\_input} gives $\ell{=}2$ as it contains two primitive actions (Figure \ref{fig:concept}b). Let $\mathrm{succ}(uv)$ be
the fraction of $f_{uv}$ that lie in successful episodes. We
define \xscore{} of a candidate as
\begin{equation}
\xscoresym(u,v) \;=\;
\underbrace{f_{uv}}_{\textstyle\text{recurrence}} \;\cdot\;
\underbrace{(\ell_u + \ell_v)^{p_\ell}}_{\textstyle\text{length}} \;\cdot\;
\underbrace{(\mathrm{succ}(uv) + \epsilon)^{p_s}}_{\textstyle\text{success}} \, ,
\label{eq:merge}
\end{equation}
and recursively merge the pair with the highest \xscore{}.
$p_\ell$ and $p_s$ set how strongly length and success weigh against recurrence (set to $1$), and $\epsilon$ is a smoothing factor on the success rate. Therefore, a long, reliable, recurrent structure outranks a frequent but short or failure-prone pair, a distinction a text tokenizer
cannot make.

A merge must also compress the corpus: it removes $f_{uv}-1$ symbols and adds
an entry of $\ell_u{+}\ell_v$ at a cost of $\eta$, a hyper-parameter keeping the corpus from collapsing into a
few giant symbols. We merge only when $f_{uv}-1 > \eta(\ell_u + \ell_v)$.
Merging ends when no pair passes it or a cap is
reached.

\textbf{What \xtree{} contains.}
The result of recursive merging yields a tree where the leaves are canonicalized actions. Each internal node is a composition of actions, a \textit{skill}, whose depth $d_v$ is one more than the deeper of its children.
The whole procedure is deterministic and auditable with zero LLM calls. Applying \xtree{} to a trajectory yields its \emph{top-level tiling}, a segmentation into maximal mined skills with uncovered positions left as primitives. 
\xtree{} is therefore a \emph{reuse prior}: it marks
which parts of a trajectory are recurring, compositional, and
success-bearing (examples in Appendix~\ref{app:casestudy}).

\subsection{\xtree{} for Offline RL: Each Node as one RL instance}
\label{sec:method-offline}

\begin{figure}
    \centering
    \includegraphics[width=\linewidth]{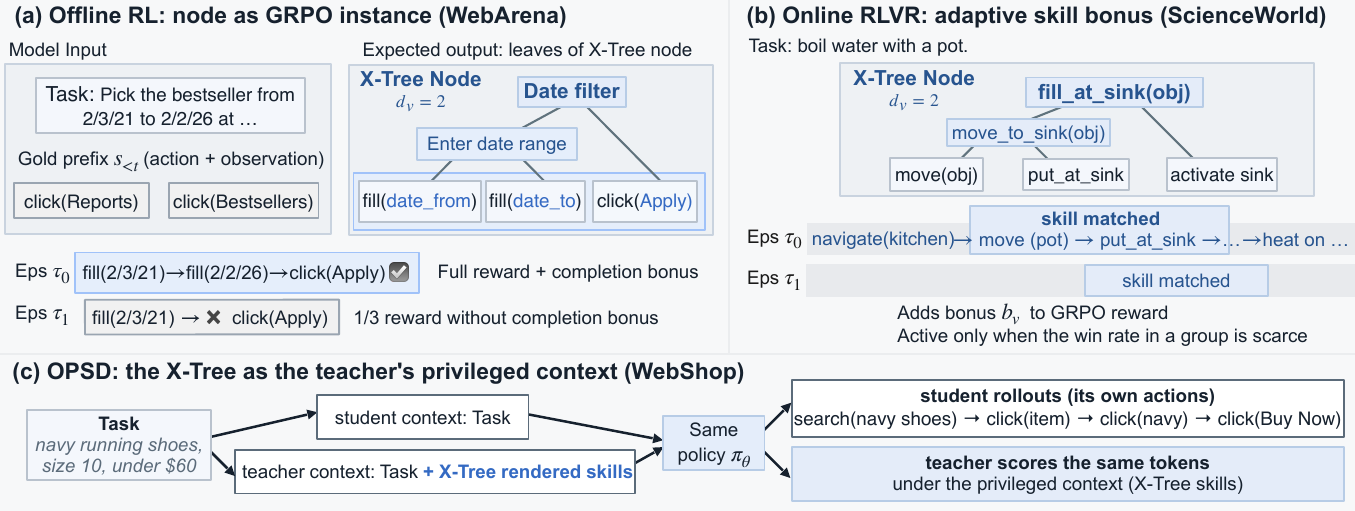}
    \caption{\textbf{Three training integrations with \xtree{}.}
(a) Offline RL: \xtree{} nodes as RL instances, partially rolled out
 from gold prefix and rewarded for node completion. (b) Online RLVR: an adaptive bonus for \xtree{} nodes a rollout executes plus outcome signal.
(c) OPSD: the same weights, given the retrieved \xtree{} skills (examples in Appendix \ref{app:opsd}) as privileged context for the self-teacher.}
\label{fig:exemplar}
\end{figure}

We have $\mathcal{D}$ without an
environment: no action can be executed.
The only available signal is agreement with the gold trajectory, which is sparse over a long horizon.
In \xtree{}, each node is a reusable 
sub-procedure which is used as one RL training instance (Figure~\ref{fig:exemplar}a):
the input is observations and actions in the trajectory before the node, the
output is actions the policy writes next (as partial rollouts), and the reward is how closely
those actions match the node. 

Formally, let $v$ be a node instantiated at position $t$ of an annotated trajectory,
$\ell_v$ its length and $d_v$ its depth (\S \ref{sec:method-tokenizer}). Conditioned on the gold actions and observations before $t$ (prefix $s_{<t}$), the policy generates at most $\ell_v$ steps. Let
$\mathrm{match}_j\in\{0,1\}$ be $1$ when step $j$ reproduces the gold
action: the same action and element. We concatenate the gold observation at step $j$ from the gold trajectory if $\mathrm{match}_j{=}1$. A rollout stops
when $\mathrm{match}_j{=}0$.
Let $c_v=\prod_{j}\mathrm{match}_j$, which is $1$ only
when every step of the node is matched. More details in Appendix  \ref{app:webarena}.
We adopt GRPO \citep{shao2024deepseekmath} with:
\begin{equation}
r \;=\; (1-\alpha)\,\underbrace{\tfrac{1}{\ell_v}\textstyle\sum_{j}\mathrm{match}_j}_{\text{step-matching reward}}
\;+\; \alpha\,\underbrace{(1+\gamma\, d_v)}_{\text{depth factor}}\,\underbrace{c_v}_{\text{node completion reward}},
\label{eq:offline-reward}
\end{equation}
with $\alpha{=}0.3$ and $\gamma{=}0.5$ as hyper-parameters. The first term is a step-matching reward for how
far the rollout stays on the gold path. The second is a bonus
for finishing the node.
In Figure~\ref{fig:exemplar}a, the \xtree{} node
\emph{date filter} fills two dates and clicks apply ($\ell_v{=}3$) with $d_v{=}2$. Rollout $\tau_0$ reproducing every action obtains full matching reward with completion bonus scaled by $d_v$. Rollout $\tau_1$ that clicks apply too early stops at the second step and gets only $1/3$ matching reward, with no completion bonus.

\subsection{\xtree{} for Online RLVR: an Adaptive Skill Bonus}
\label{sec:method-online}

With an environment and a verifier, RLVR is possible.
Early in training, groups with no successful rollout give GRPO zero advantage and hence no gradient.
\xtree{} provides the missing contrast: a rollout that carries out more mined skills is distinguished from one that does not (Figure~\ref{fig:exemplar}b).
And the signal gets down-weighted as the verifier becomes
informative on its own.

Let $\tau_i$ be the $i$-th rollout of a group $g$, $R_{\mathrm{out}}(\tau_i)$ the verifier reward, $\mathrm{matched}(\tau_i)$ the nodes in \xtree{} whose action sequence appears in
$\tau_i$ (each counted once per episode): the nodes $\tau_i$ actually
executes. Let $b_v$ be the reward an execution of node $v$ earns, specific to an environment. Let
$\mathrm{w}_g$ be the fraction of the group the verifier counts as successes, $w_{\mathrm{ref}}$ a threshold, and $\lambda_g$ the weight which the bonus carries in $g$: $\lambda_g$ equals
$\lambda_0$ when no rollout succeeds and falls to zero once ${w}_g{=}w_{\mathrm{ref}}$, so the bonus is at full strength when the
verifier cannot separate and absent on those it can. We
define reward:
\begin{equation}
r_i \;=\; \underbrace{R_{\mathrm{out}}(\tau_i)}_{\text{verifier}}
\;+\; \lambda_g \!\!\sum_{v \in \mathrm{matched}(\tau_i)}\!\! b_v,
\qquad
\lambda_g \;=\; \lambda_0\,\mathrm{clip}\!\Big(1 - \tfrac{{w}_g}{w_{\mathrm{ref}}},\,0,\,1\Big).
\label{eq:online-reward}
\end{equation}
$b_v$ stands for the value of the \xtree{} node $v$, \textit{e.g.,} depth $d_v$, where a deeper node is a longer and harder composition.
In Figure~\ref{fig:exemplar}b, \texttt{fill\_at\_sink(obj)} is a depth-2
node. A rollout $\tau_0$ that executes it earns the bonus 
$b_\text{fill\_at\_sink}$.
\S\ref{sec:setup} specifies $b_v$ designs in each
environment.
Appendix~\ref{app:rewards} traces
$\lambda_g$ over training step: an adaptive device with an operating window, not a universal add-on.

\subsection{\xtree{} for On-policy Self-Distillation: a Privileged Context}
\label{sec:method-opsd}

OPSD trains a model on its own rollouts toward a
teacher that is the \emph{same} model conditioned on \textit{privileged information}
\citep{zhao2026self, shenfeld2026self}. A per-token distillation term is added
to the RL objective, so the student receives dense token-level guidance. Hinted by Table \ref{tab:prompted}, \xtree{} as advice in context can enhance a model's multi-step reasoning.
We therefore add \xtree{} to the
teacher as privileged context in OPSD.
Since teacher and student are the same policy differing only in context, any gap between them is attributable to the context (LLM-written or \xtree{} skills).

Let $y=(y_1,\dots,y_{|y|})$ be the response the student sampled for a prompt $s^{\mathrm{stu}}$, and $s^{\mathrm{tea}}$ the same prompt with retrieved \xtree{} skills prepended: the response tokens are identical under
both, so log-probabilities align position for position. Write $\bar{\theta}$ for
the weights held fixed for the teacher's forward pass, $\delta_t = \log
\pi_{\bar{\theta}}(y_t \mid s^{\mathrm{tea}}) - \log \pi_{\theta}(y_t \mid
s^{\mathrm{stu}})$ for the per-token gap, positive exactly on the
tokens the privileged context makes more likely. The gap also decides how much
of the teacher each token copies: $g_t = \sigma(\beta\,\delta_t)$, with
$\sigma$ the logistic function and $\beta$ the sharpness, is a confidence gate. It is near $1$ where the teacher leads and near $0$ where the student is already the
more confident, so the student is pulled toward the teacher only on
the tokens the skills inform. A coefficient $c$ sets the weight of this
term against the RL loss (set to $0.01$).
The objective adds one term to the RL loss:
\begin{equation}
\mathcal{L} \;=\; \mathcal{L}_{\mathrm{GRPO}}(R_{\mathrm{out}})
\;+\; c\,\underbrace{\tfrac{1}{|y|}\textstyle\sum_{t} g_t\,
\big(\log \pi_{\bar{\theta}}(y_t \mid s^{\mathrm{tea}}) - \log \pi_{\theta}(y_t \mid s^{\mathrm{stu}})\big)}_{\text{per-token distillation}}.
\label{eq:opsd}
\end{equation}
Minimising it raises the student probability that the
privileged context finds likely. The rendered \xtree{} writes each node as one line of text with its success rate and occurrence count (Appendix~\ref{app:render}). 

\section{Experimental setup}
\label{sec:setup}

\textbf{Offline RL: WebArena \citep{zhou2023webarena}.}
We mine 256 \xtree{} skills from 7.9k trajectories \citep{gandhi2025gobrowse}.
We use Qwen2.5-7B-Instruct with SFT on half the pool, then
one GRPO stage on \xtree{} nodes from the other half by verl. Appendix~\ref{app:hyper} lists training details.
Evaluation runs the 694 deterministic (non-fuzzy) tasks, 30-step cap with official reward.
We evaluate success rate (SR) with normalization, \textit{i.e.,} a task counts as
success if the official evaluator returns $1$, or, for the {exact match}
tasks, if the normalized gold string appears verbatim in the
final output. All baselines are based on the same harness. Appendix~\ref{app:webarena} shows more details.

\textbf{Online RLVR: ScienceWorld \citep{wang2022scienceworld} and WebShop \citep{yao2022webshop}, three scales.}
We use SFT warm starts from 200 (ScienceWorld) and 500 (WebShop) trajectories at 1.5B/3B/7B, then a GRPO stage.
The bonus (Eq.~\ref{eq:online-reward}) uses $\lambda_0{=}0.75$ and $w_{\mathrm{ref}}{=}0.4$ on ScienceWorld, $w_{\mathrm{ref}}{=}0.85$ on WebShop (training details in Appendix~\ref{app:hyper}, hyperparameter sensitivity in Appendix \ref{app:lambda_sens}).

\textbf{{i)}} ScienceWorld: we mine 80 \xtree{} skills  from 1{,}673 trajectories \citep{xi2024agentgym}. 
We evaluate success rate (score${=}100$) on three generalization folds \citep{zhang2025rlvmr}: G0 (seen tasks, $n{=}1.6k$), G1 (unseen variations, $n{=}1.6k$), G2 (unseen tasks, $n{=}0.5k$, trained on 19 types and evaluated on the remaining 10). 
A matched skill node earns $b_v = 1+\gamma\,d_v$, granted once per episode. When two matched skills nest, the outer one is rewarded the difference between its $b_v$ and the largest already rewarded inside it, so a nested family contributes what its largest member is worth rather than the sum.
\textbf{{ii)}} WebShop: we mine 48 skills from 1{,}824 trajectories \citep{song2024eto}.
We evaluate two metrics on 512 held-out episodes: \emph{success}, the fraction of purchases that meet every requirement of the task goal, and \emph{graded score}, the partial reward in $[0,1]$ for the requirements met.
A matched \xtree{} skill node earns $b_v = s_v k_v / T$: its corpus success rate $s_v$ times the number $k_v$ of steps in the \xtree{} node that help the task goal, over the episode length $T$.

\textbf{OPSD: ScienceWorld and WebShop, three scales.}
OPSD shares the \xtree{}, the SFT warm start and evaluation with RLVR for both environments.
It changes only the skill bank the self-teacher reads: LLM-written or the \xtree{} rendering.
\textbf{{i)}} ScienceWorld: we use our implementation of Eq.~\ref{eq:opsd}. The teacher reads one \xtree{} rendering per task family, retrieved by the task line: the most relevant skill node to the task family. The LLM-written bank is by gpt-oss-120b from the same corpus with the same retrieval keys.
\textbf{ii)} WebShop: we use the released trainer \citep{lu2026sdar}. The LLM-written bank is by GPT-o3. \xtree{} rendered examples are in Appendix \ref{app:render}.

We use Qwen2.5-Instruct and report the mean over three seeds. Appendix~\ref{app:stability} shows stability analysis.

\section{Results and Analysis}
\label{sec:results}
\label{sec:analysis}

We validate the three training integrations (\S\ref{sec:method}) over three environments and three model scales. We show the main result first and then the ablations that
isolate where its gain comes from.

\subsection{Offline RL on WebArena}
\label{sec:webarena}

\begin{table}[t]
\definecolor{tOne}{HTML}{F9FAFB}\definecolor{tTwo}{HTML}{EEF4FB}\definecolor{tThree}{HTML}{DCE8F7}
\centering
\caption{\textbf{Per-site SR (\%) on WebArena (Qwen2.5-7B)}. Go-Browse is our reproduced result. Every RL starts from SFT on half the pool. ``SFT$+$offlineRL'' rows keep the same offline RL stage: one rewards only the outcome of a whole trajectory, the other replaces \xtree{} nodes by random spans matched in node count and length. ``Plain mixing'' trains on all \xtree{} nodes at once without completion reward. ``Curriculum'' trains in order of depth. ``\xtree{} Full'' is our recipe as designed in Eq.~\ref{eq:offline-reward}. Our \xtree{} with offline RL consistently outperforms other recipes. \textbf{Bold}: best per column.}
\label{tab:webarena-main}
\small
\setlength{\tabcolsep}{4pt}
\setlength{\aboverulesep}{0pt}\setlength{\belowrulesep}{0pt}
\renewcommand{\arraystretch}{1.18}
\resizebox{\textwidth}{!}{%
\begin{tabular}{lcccccc|c}
\toprule
\textbf{Method} & \textbf{gitlab (187)} & \textbf{shopping (158)} & \textbf{admin (156)} & \textbf{reddit (112)} & \textbf{map (67)} & \textbf{wiki (14)} & \textbf{All (694)} \\
\midrule
\rowcolor{tOne}\multicolumn{8}{c}{\emph{{w/o RL}}} \\
\rowcolor{tOne}Qwen2.5-7B-Instruct (base) & 6.4 & 7.8 & 7.7 & 2.4 & 9.0 & 2.4 & 6.5 \\
\rowcolor{tOne}Go-Browse (SFT-Full w/ 2 epochs) & 16.8 & 21.7 & 22.9 & 16.4 & 10.4 & 7.1 & 18.4 \\
\rowcolor{tOne}\quad w/ 4 epochs (matched compute) & 17.5 & 19.4 & 24.4 & 16.1 & 13.4 & 11.9 & 18.8 \\
\midrule
\rowcolor{tTwo}\multicolumn{8}{c}{\emph{{SFT $+$ offline RL (Ours)}}} \\
\rowcolor{tTwo}Offline RL w/ Outcome reward & 15.5 & 23.4 & 23.1 & 15.2 & 7.5 & 14.3 & 18.2 \\
\rowcolor{tTwo}Offline RL w/ Random span & 18.4 & 20.9 & 26.6 & 14.7 & 11.9 & 14.3 & 19.5 \\
\midrule
\rowcolor{tThree}\multicolumn{8}{c}{\emph{{SFT $+$ offline RL (Ours) $+$ \xtree{} (Ours)}}} \\
\rowcolor{tThree}\xtree{} Plain mixing, all depths & 16.1 & 24.5 & 25.4 & 19.9 & \textbf{14.4} & \textbf{23.8} & 20.7 \\
\rowcolor{tThree}\xtree{} Curriculum, $d_v{=}1{\to}2...$ & 17.8 & 21.5 & 28.6 & \textbf{20.8} & 12.4 & \textbf{23.8} & 21.2 \\
\rowcolor{tThree}\textbf{\xtree{} Full} & \textbf{22.0} & \textbf{26.6} & \textbf{29.3} & 18.8 & 10.4 & 14.3 & \textbf{22.9} \\
\bottomrule
\end{tabular}}
\vspace{-10pt}
\end{table}

\paragraph{\xtree{} turns a fixed trajectory pool into $+4.5$ SR without any other information.}
With the same trajectories and no environment at training, offline RL with \xtree{} reaches a $24\%$ relative gain over Go-Browse recipe (SFT-Full) (Table~\ref{tab:webarena-main}, $22.9$ \textit{vs} $18.4$). 
Matching RL compute with two extra SFT epochs raises SFT-Full by only $0.4$, so the gain is not from compute.
Our offline RL approach reliably boosts SFT performance, and adding \xtree{} leads to further gains. This validates both our RL method and \xtree{} structure, significantly improving the multi-turn agent learning under the same budget.
Our improvement is consistent across all sites, with especially large margins on the three biggest sites compared to SFT recipe: admin improves the most ($+6.4$ SR), followed by gitlab, the most procedure-intensive site ($+5.2$ SR), and then shopping ($+4.9$ SR). The margin shrinks on reddit, map and wiki, since they are driven mainly by query formulation, not procedural execution. Moreover, the hierarchical reward we propose (\S \ref{sec:method-offline}) performs better than simply mixing \xtree{} nodes with a flat reward or a depth-based curriculum training. This suggests the presence of reusable structure in trajectory data that merits further investigation. On the same budget, \xtree{} captures informative signals that facilitate enhanced compositional generalization in offline RL.
{Our recipe also transfers to another model family (GLM-4) and larger scale (14B) (Appendix \ref{app:webarena}).}

\subsubsection{Which part of the recipe carries the gain, the tree or the RL method?}
\label{sec:ablation}
Our recipe changes two things compared to Go-Browse SFT on the same corpus: the \emph{training instance} (\xtree{} nodes instead of whole trajectories) and a \emph{novel RL method} (with a step-matching reward and a completion bonus). Table~\ref{tab:webarena-ablations} ablates one at a time against the full recipe.

\begin{wraptable}{r}{0.36\textwidth}
\vspace{-20pt}
\centering
\small
\setlength{\tabcolsep}{2.5pt}
\renewcommand{\arraystretch}{0.95}
\caption{{Ablations} on WebArena: \emph{Tree Structure Validation} replaces \xtree{}  with other settings fixed; \emph{Offline RL Validation} replaces the training method on identical data.}
\label{tab:webarena-ablations}
\begin{tabular}{@{}lcc@{}}
\toprule
\textbf{Method} & \textbf{SR} & \textbf{$\Delta$} \\
\midrule
\textbf{\xtree{} Full} & \textbf{22.9} & -- \\
\midrule
\multicolumn{3}{@{}l}{\emph{\textbf{Tree Structure Validation}}} \\
 Whole trajectory (no tree) & 19.9 & $-3.0$ \\
 Random span   & 19.5 & $-3.4$ \\
 Random tree  & 17.9 & $-5.0$ \\
\midrule
\multicolumn{3}{@{}l}{\emph{\textbf{Offline RL Validation}}} \\
\xtree{} Plain mixing & 20.7 & $-2.2$ \\
 \xtree{} Curr. $d_v{:}1{\to}2...$ & 21.2 & $-1.7$\\
 Binary outcome reward & 18.2 & $-4.7$ \\
 SFT-Full, 2 epochs & 18.4 & $-4.5$ \\
 SFT-Full, 4 epochs & 18.8 & $-4.1$ \\
\bottomrule
\end{tabular}
\vspace{-10pt}
\end{wraptable}

\emph{\textbf{Against \xtree{}.}} We test whether our \xtree{} structure matters. Training on the whole trajectory with the same RL stage drops $3.0$. Random span matches the count and length histogram of \xtree{}, which keeps the chunking but moves the boundaries, which drops $3.4$. A random tree with the same number of merges, which replaces \xscore{} by random, drops $5.0$, below SFT-only ($18.4$).
This indicates that the reusable structure built by \xscore{} provides useful training signal.

\emph{\textbf{Against the RL method.}} We vary training on identical data (the \xtree{} nodes). SFT-full is significantly behind our RL even without an environment. \emph{Plain mixing} trains without completion bonus in Eq \ref{eq:offline-reward}, which drops $2.2$. \emph{Curriculum} also removes the completion term and trains bottom-up by $d_v{=}1$ then $2$, ..., which drops $1.7$. These two variants show that \xtree{} entering the reward helps harder compositions, whereas a flat reward or data curriculum under-credits reusable structures. Moreover, with \xtree{} these two variants perform well on sites requiring query formulation, \textit{i.e.,} reddit, map and wiki, while our step-matching reward and node completion bonus scaled by depth help more with procedural execution, gitlab, shopping and admin (Table \ref{tab:webarena-main}).
A binary trajectory-level reward drops $4.7$, showing that the outcome signal is too sparse to learn the multi-turn reasoning. This suggests that our offline RL integration with \xtree{} helps the consolidation of the reusable structure. 

\textbf{Takeaway.} The gain comes from both our structure and method: the \xtree{} node captures the composition of reusable skills which guides training with step-matching and depth-scaled reward.

\subsection{Online RLVR on ScienceWorld and WebShop}
\label{sec:online}
\label{sec:sciworld}

\begin{figure}[t]
\centering
\includegraphics[width=\textwidth]{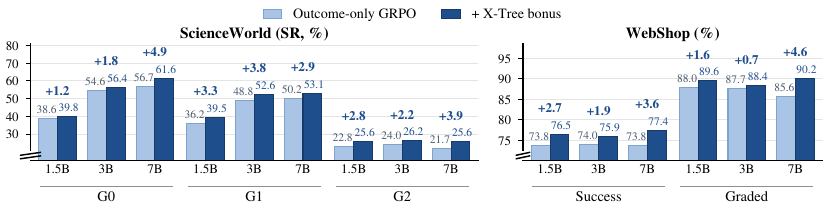}
\vspace{-20pt}
\caption{\textbf{Online RLVR results.} ScienceWorld SR on three
generalization levels \citep{zhang2025rlvmr}; WebShop success and graded score. \textbf{Bold}: the difference of a pair. The \xtree{} integration consistently outperforms outcome-only GRPO.
Statistical significance in Appendix~\ref{app:stability}.}
\label{fig:online-rlvr}
\vspace{-5pt}
\end{figure}

Figure~\ref{fig:online-rlvr} adds the adaptive \xtree{} bonus
(\S\ref{sec:method-online}) to outcome-only GRPO at 1.5B, 3B and 7B, from a
SFT warm start on ScienceWorld and WebShop. Every pair of bars is a comparison that changes only the reward.
Refer to Appendix~\ref{app:sciworld} for further analysis. Statistical significance is in Appendix~\ref{app:stability}.

\textbf{ScienceWorld: \xtree{} is ahead consistently and stably, with larger margin at larger scales.}
On seen tasks (G0, G1), \xtree{} integration is above outcome-only GRPO in all comparisons and scales well with model sizes (by up to $4.9\%$ SR). 
On held-out tasks (G2), \xtree{} leads at every scale, by up to $3.9\%$ SR. This indicates that \xtree{} introduces reusable experience that can be captured by the policy through reward signal. Such structured experience captures how a recurrent and success-bearing skill is composed from sub-skills, which can transfer to tasks the model never trained on.
We find that \xtree{} is robust with a mining corpus excluding the held-out tasks (Appendix~\ref{app:leakage}).

\textbf{WebShop: \xtree{} is superior at every scale, consistently and stably.}
\label{sec:webshop}
The adaptive \xtree{} bonus improves success (by up to $3.6\%$) and graded score (by up to $4.6\%$). The gain scales well across model sizes (7B gap is the largest). 
This shows that the reusable structure in  \xtree{} and our designed training signal enhances the overall performance by facilitating multi-turn procedure progress.

\subsubsection{When does the \xtree{} adaptive bonus help?}
\label{sec:bonus-window}
The \xtree{} integration with RLVR (Eq.~\ref{eq:online-reward}) is designed to build reusable structure when the verifier signal is scarce (\textit{e.g.,} early training stage or small rollout $n$), then gradually fade when it is dense. We test when the adaptive bonus truly helps by varying $rollout.n$ on ScienceWorld (Table~\ref{tab:rollout-n}).

\begin{wraptable}{r}{0.38\textwidth}
\vspace{-21pt}
\centering
\small
\setlength{\tabcolsep}{3pt}
\renewcommand{\arraystretch}{0.98}
\caption{{Rollout number $n$ scaling (ScienceWorld, 1.5B)} on three levels.}
\label{tab:rollout-n}
\begin{tabular}{@{}rlccc@{}}
\toprule
\textbf{$n$} & \textbf{Setting} & \textbf{G0} & \textbf{G1} & \textbf{G2} \\
\midrule
2 & outcome-only & 12.9 & 13.7 & 11.8 \\
  & + \xtree{} bonus & \textbf{14.6} & \textbf{15.5} & \textbf{13.0} \\
\midrule
4 & outcome-only & 24.2 & 23.9 & 12.1 \\
  & + \xtree{} bonus & \textbf{28.0} & \textbf{27.5} & \textbf{16.5} \\
\midrule
8 & outcome-only & 38.6 & 36.2 & 22.8 \\
  & + \xtree{} bonus & \textbf{39.8} & \textbf{39.5} & \textbf{25.6} \\
\midrule
16 & outcome-only & 47.5 & \textbf{49.0} & 26.6 \\
   & + \xtree{} bonus & \textbf{51.0} & 47.9 & \textbf{27.4} \\
\bottomrule
\end{tabular}
\vspace{-8pt}
\end{wraptable}

Our recipe is ahead in $11{/}12$ cells.
The gain is larger at smaller rollout size (from $3.6$ to $4.4$ at $n{=}4$), where outcome-only GRPO receives less reward signal. And the lead narrows at larger $n$, with potentially denser reward signal, where the adaptive bonus is designed to anneal out. This confirms that \xtree{} helps with sparse reward and defers to outcome reward when the verifier signal is dense.

Three mechanisms (Eq.~\ref{eq:online-reward}) explain the window. \textbf{i)} \emph{Weight resolution}: $\lambda_g$ depends on win rate among $n$ rollouts, taking values in $\{0,\tfrac12,1\}$ at $n{=}2$. \textbf{ii)} \emph{The gradient}: in an all-fail group the outcome terms cancel and the in-group gradient is driven by \xtree{} node matching alone, keeping the training alive. \textbf{iii)} \emph{Anneal versus learning speed}: the measured late-training $\lambda_g$ exposure falls from $0.50$ at $n{=}2$ to $0.18$--$0.26$ at $n{=}16$, so the bonus is active while the policy is weak
and fades once it is strong. Training dynamic of $\lambda_g$ is in Appendix \ref{app:rewards}. 

\textbf{Takeaway.} The adaptive bonus
is active when the verifier is silent (\textit{e.g.,} early training stage or a small $n$) and fades when it is dense, which builds reusable structure to facilitate multi-step reasoning.

\subsubsection{Is the gain from the extra data used to mine \xtree{}?}
\label{sec:webshop-signal}
\begin{wraptable}{r}{0.37\textwidth}
\vspace{-22pt}
\centering
\small
\setlength{\tabcolsep}{2.5pt}
\caption{{Resource-matched runs (WebShop, 1.5B).} All rows
use same resources and differ only in reward.}
\label{tab:webshop-controls}
\vspace{1pt}
\begin{tabular}{@{}lcc@{}}
\toprule
\textbf{Warm start} & \textbf{outcome} & \textbf{$+$\xtree{} ($\Delta$)} \\
\midrule
500 Samples & 73.8 & \textbf{74.9} $(+1.1)$ \\
1{,}016 Success & 74.9 & \textbf{78.1} $(+3.2)$ \\
1{,}824 All & 76.4 & \textbf{77.4} $(+1.0)$ \\
\bottomrule
\end{tabular}
\vspace{-12pt}
\end{wraptable}

In the main results, all experiments are trained by SFT and RLVR with identical data, whereas \xtree{} is mined from a trajectory corpus. One question is whether \xtree{}'s gain comes from the mining corpus.
We conduct resource-matched experiments (WebShop, 1.5B, Table~\ref{tab:webshop-controls}). We  mine \xtree{} from the SFT data and vary the data volume, so that all resources are the identical.
We find that our \xtree{} still manages to capture reusable structures to better guide training with exactly the same budget.

\begin{wraptable}{r}{0.37\textwidth}
\vspace{-10pt}
\centering
\small
\setlength{\tabcolsep}{2.5pt}
\caption{{Mining-corpus size (WebShop, 1.5B).} \xtree{} is mined from $N$ random trajectories of the corpus.}
\label{tab:webshop-variants}
\vspace{1pt}
\begin{tabular}{@{}rcc@{}}
\toprule
\textbf{\# Traj} & \textbf{Skills} & \textbf{Success (\%)} \\
\midrule
50 & 9 & 68.7 \\
100 & 11 & 75.2 \\
200 & 16 & 72.9 \\
500 & 24 & 74.4 \\
1{,}824 & 48 & \textbf{76.5} \\
\bottomrule
\end{tabular}
\vspace{-10pt}
\end{wraptable}

\textbf{Mining \xtree{} does not require many trajectories.} We re-mine \xtree{} from
random subsets of the corpus, with other settings (\textit{e.g.,} data, training) fixed (Table~\ref{tab:webshop-variants}). The 100-trajectory \xtree{} is already above outcome reward with 500 sample SFT warm start ($75.2$ vs.\ $73.8$). Success reaches $76.5$ with full corpus, not monotonically in between, showing the robustness and efficiency of building \xtree{}.

\textbf{Takeaway.} \xtree{} wins by the structure and reward design. The hierarchy, which captures how a reusable skill is composed from sub-skills, facilitates generalization under the same budget. Mining \xtree{} structure does not require a large trajectory pool.

\subsection{OPSD on ScienceWorld and WebShop}
\label{sec:opsd-results}

\begin{figure}[t]
\centering
\includegraphics[width=\textwidth]{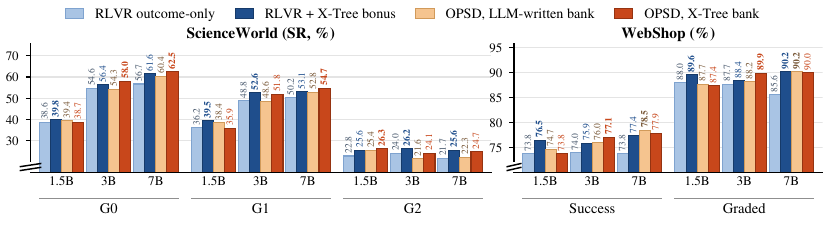}
\vspace{-20pt}
\caption{\textbf{Validating OPSD with \xtree{}.} ScienceWorld SR on three levels \citep{zhang2025rlvmr}; WebShop success and graded score.
OPSD with \xtree{} changes only the privileged context of the self-teacher. \xtree{} is comparable with LLM-written bank. Statistical significance in Appendix~\ref{app:stability}.}
\label{fig:webshop-opsd}
\vspace{-5pt}
\end{figure}

Figure~\ref{fig:webshop-opsd} validates OPSD with \xtree{} against the online RLVR on ScienceWorld and WebShop at 1.5B, 3B and 7B. RLVR and OPSD integrate the same \xtree{} in two ways, as a reward bonus or as the privileged context of the self-teacher. Within OPSD only the skill bank changes, LLM-written (gpt-oss-120b on ScienceWorld, GPT-o3 on WebShop) or the rendered \xtree{}.

\paragraph{\xtree{} performs on par with the LLM-written skill bank at zero LLM cost.}
Across both environments, different generalization levels and three model scales, our rendered \xtree{} is comparable to the LLM-written bank, ahead at most of the runs (by up to $+3.7$ on ScienceWorld and $+1.7$ on WebShop).
This indicates that \xtree{}, mined deterministically with no LLM in the loop under the same budget, is as good to the self-teacher as an LLM-written artifact for the same environment.

\paragraph{\xtree{} provides useful signals for OPSD, with larger models pushing better performance over RLVR.} 
OPSD with our \xtree{} is above outcome-only RLVR in most cells, and the margin grows with the model scale. At 7B, OPSD reaches $+5.8$ SR on ScienceWorld and $+4.4$ graded score on WebShop, while at 1.5B and 3B the gains stay within a few points. OPSD distills from the model's own predictions under the skill text, so its gain depends on how well the base model reads and follows that text. Compared to the \xtree{} bonus in RLVR, OPSD is ahead at 7B on ScienceWorld seen tasks (G0, G1) and comparable at smaller scales, while RLVR is ahead of outcome-only throughout. This indicates that the two integrations of \xtree{} are both effective ways of incorporating the reusable structure: \xtree{} enters RLVR as a reward signal which helps from smaller to larger models; while \xtree{}, acting as privileged context in OPSD, further improves the performance once the model can well exploit the skill text. We ablate the skill bank in Appendix~\ref{app:opsd}.

\section{Conclusion}
\label{sec:conclusion}

Limited experience is the current bottleneck in training multi-step agents.
Standard recipes consume the action sequence uniformly and bottom-up, without the hierarchy that lets human learners generalize more efficiently with the same data.
We tokenize reusable experience to recover that hierarchy from a trajectory pool as \xtree{}, with zero LLM calls. Each \xtree{} node captures how a recurrent and success-bearing skill is composed from sub-procedures, guiding efficient generalization.
\xtree{} enters training in three settings: as the training instance in offline RL, as an adaptive bonus that keeps online RLVR ahead while the outcome signal is scarce, and as the privileged context in OPSD where it matches the LLM-written skill bank.
Analyses attribute the gain to the \xtree{} structure and our integration design.
These results hold across three environments and three model scales.

\textbf{Limitations.}
We mine \xtree{} once from a fixed pool, and integrate \xtree{} with training without improving it.
Mining from the policy's own trajectories would potentially let the structure and the policy improve together in a loop.
We evaluate the three integrations one per setting and never combine them. A possible future work is to train a single model on \xtree{} nodes, taking the \xtree{} bonus, and distill from \xtree{} at once.
\xtree{} also only reorganizes experience that already exists.
Because it records which pairs recur and which succeed, it marks where a corpus is thin, identifying the trajectories worth synthesizing to fill the gap.

\subsubsection*{AI use statement}
We used an AI assistant (Claude Code) in two roles. For writing, it
polished text drafted by the authors and typeset tables from the result files the authors provided. The authors reviewed
and edited all of its output. For retrieval and discovery, it helped locate
related work and verify bibliographic details, which the authors read and
selected. The research questions, methods, experiments and conclusions are the authors' own, and the authors take full responsibility for the content.

\subsubsection*{Ethics statement}
This work uses publicly available benchmarks and simulated environments. It does not involve
human-subject experiments, collection of personally identifying information, or deployment that affects individuals. We follow the ICLR Code of Ethics and report the methods and
limitations of the study transparently.

\subsubsection*{Reproducibility statement}
The canonicalizers, miner and training configurations and the evaluation harness will be released with the paper. The
evaluation protocols are stated in \S\ref{sec:setup}. The spread behind every cell of the online tables is in Appendix~\ref{app:stability}.

\bibliography{references}
\bibliographystyle{iclr2027_conference}

\appendix

\clearpage
\section{Implementation details}
\label{app:hyper}
\suppressfloats[t]

\subsection{Training details}
\begin{table}[t]
\centering
\footnotesize
\setlength{\tabcolsep}{4pt}
\renewcommand{\arraystretch}{0.96}
\caption{\textbf{\xtree{} construction and SFT hyperparameters.} The canonicalization and merging are per environment. Merging stops at the compression constraint 
\S\ref{sec:method-tokenizer} or the skill cap.}
\label{tab:hyper-mining}
\begin{tabular}{@{}llll@{}}
\toprule
\textbf{Setting} & \textbf{WebArena} & \textbf{ScienceWorld} & \textbf{WebShop} \\
\midrule
\multicolumn{4}{@{}l}{\emph{\textbf{\xtree{} Mining} (Eq.~\ref{eq:merge})}} \\
$p_\ell$, $p_s$ & 1, 1 & 1, 1 & 1, 1 \\
Success floor $\epsilon$ & 0.001 & 0.001 & 0.001 \\
$\eta$ & 0.3 & 0.5 & 0.5 \\
Skill cap & 256 & 80 & 60 \\
Max.\ skill length (actions) & 12 & 8 & 8 \\
Min.\ pair frequency & 4 & 5 & 5 \\
Min.\ skill success rate & 0 & 0.9 & 0 \\
Min.\ task types per skill & 1 & 2 & 1 \\
Mining corpus (trajectories) & 7{,}974 & 1{,}673 & 1{,}824 \\
Primitive action types & 139 & 84 & 17 \\
Skills mined (bound by) & 256 (cap) & 80 (cap) & 48 (compression) \\
\midrule
\multicolumn{4}{@{}l}{\emph{\textbf{SFT warm start}}} \\
Epochs & 2 & 5 & 5 \\
Learning rate & $2{\times}10^{-5}$ & $1{\times}10^{-5}$ & $1{\times}10^{-5}$ \\
Global batch (sequences) & 4 & 256 & 256 \\
Max.\ sequence (tokens) & 24{,}000 & 4{,}096 & 8{,}192 \\
\bottomrule
\end{tabular}
\end{table}

\begin{table}[t]
\centering
\small
\setlength{\tabcolsep}{4pt}
\renewcommand{\arraystretch}{0.96}
\vspace{-15pt}
\caption{\textbf{RL hyperparameters for the three environments.} }
\label{tab:hyper-rl}
\begin{tabular}{@{}llll@{}}
\toprule
\textbf{Setting} & \textbf{WebArena (offline)} & \textbf{ScienceWorld} & \textbf{WebShop} \\
\midrule
Rollouts per group $n$ & 2 & 8 & 8 \\
Optimizer steps & 805 & 100 & 150 \\
Units per step & 6 nodes & 8 tasks & 16 tasks \\
Learning rate & $5{\times}10^{-7}$ & $1{\times}10^{-6}$ & $1{\times}10^{-6}$ \\
KL to the warm start & none & 0.01 & 0.01 \\
Mini-batch & 2 & 256 & 64 \\
Max.\ prompt (tokens) & 16{,}384 & 6{,}000 & 4{,}096 \\
Max.\ response (tokens) & 16{,}384 & 1{,}024 & 512 \\
Temperature (train / eval) & 0.7 / 0 & 1.0 / 0.4 & 1.0 / 0.4 \\
Episode cap (env.\ steps) & node horizon & 30 & 15 \\
Invalid-action penalty & -- & 0.1 & 0.1 \\
\midrule
\multicolumn{4}{@{}l}{\emph{\textbf{Reward constants}}} \\
$\alpha$, $\gamma$ (Eq.~\ref{eq:offline-reward}) & 0.3, 0.5 & -- & -- \\
$\lambda_0$, $w_{\mathrm{ref}}$ (Eq.~\ref{eq:online-reward}) & -- & 0.75, 0.4 & 0.75, 0.85 \\
Node bonus $b_v$ & -- & $1+0.5\,d_v$ & $s_v k_v / T$, ${\times}10$ \\
Self-distillation coefficient $c$ (OPSD) & -- & 0.01 & 0.01 \\
Gate sharpness $\beta$ (Eq.~\ref{eq:opsd}) & -- & 5 & 0 \\
\bottomrule
\end{tabular}
\vspace{-12pt}
\end{table}

\begin{table}[t]
\centering
\small
\setlength{\tabcolsep}{4pt}
\renewcommand{\arraystretch}{0.96}
\caption{\textbf{\xtree{} construction and deployment for the prompted-agents setting} (Table~\ref{tab:prompted}). Merging follows Eq.~\ref{eq:merge} and stops at the compression constraint or the skill cap.}
\label{tab:hyper-prompted}
\begin{tabular}{@{}>{\raggedright\arraybackslash}p{3.3cm}>{\raggedright\arraybackslash}p{10.4cm}@{}}
\toprule
\textbf{Setting} & \textbf{MiniWoB++} \\
\midrule
Backbones & Qwen2.5-7B, Qwen3-8B, Qwen3-14B \\
Mining corpus & episodes of the evaluated backbone on the 53 training task families, 40 per family \\
Trajectories (successful) & 2{,}125 (886) / 2{,}120 (1{,}116) / 2{,}120 (1{,}239) \\
Filtering & none, failures kept with their label \\
Canonical action & \ctok{verb}{role}, role from the DOM  (Table~\ref{tab:canon-miniwob-mind2web}) \\
$p_\ell$, $p_s$ & 1, 1.5 \\
Success floor $\epsilon$ & 0.001 \\
$\eta$ & 0.3 \\
Skill cap & 200 \\
Max.\ node length & 10 \\
Min.\ pair frequency & 4 \\
Min.\ node success rate & 0 \\
Skills mined (bound by) & 90 / 71 / 60 (compression) \\
Deployment & advisory hint per step, top-8 skills by utterance $n$-gram \\
Evaluation & 42 task families (26 seen, 16 unseen) $\times$ 20 seeds $=$ 840 episodes \\
\bottomrule
\end{tabular}
\end{table}

Tables~\ref{tab:hyper-mining} to~\ref{tab:hyper-prompted} show the hyperparameters behind experiments. Table~\ref{tab:hyper-prompted} covers the \xtree{} in-context evaluation in Appendix~\ref{app:prompted}.
The offline stage is our node-as-instance offline RL based on 
verl. ScienceWorld runs with the code of 
\cite{zhang2025rlvmr} with our reward bonus. WebShop's RLVR is based on
verl-agent \citet{feng2025gigpo}, and the OPSD is based on the code of \citep{lu2026sdar}.

Hinted by Table \ref{tab:webshop-variants}, building \xtree{} does not need a large trajectory pool to help. However, previous studies \citep{cheng2025atomic} show that RL generalization requires a decent SFT-built base. Every online run in this paper therefore starts from a small warm start: on ScienceWorld and WebShop we sample 200 and 500 trajectories from the trajectory pool. Within every
comparison we report, all experiments (including RLVR and OPSD) start from the same warm-start checkpoint for a model scale in an environment.

\subsection{Canonicalization examples}
\label{app:canon}
\begin{table}[t]
\centering
\small
\setlength{\tabcolsep}{4pt}
\renewcommand{\arraystretch}{0.95}
\vspace{-5pt}
\caption{\textbf{Canonicalization on WebArena.} A BrowserGym action string
to \ctok{verb}{role}: verb is from the function name, role from
the accessibility-tree role of the element the action addresses.}
\label{tab:canon-webarena}
\begin{tabular}{@{}p{7.5cm}p{2.4cm}p{3.5cm}@{}}
\toprule
\textbf{Raw action} & \textbf{Canonical} & \textbf{Kept in slots} \\
\midrule
\multicolumn{3}{@{}l}{\emph{\textbf{Verb}, from the BrowserGym function name}} \\
\texttt{click(bid)}, \texttt{dblclick(bid)} & \texttt{click} & element id \\
\texttt{fill(bid,\,text)}, \texttt{type(bid,\,text)} & \texttt{type} & element id, typed text \\
\texttt{select\_option(bid,\,opt)} & \texttt{select} & element id, option value \\
\texttt{press(bid,\,key)}, \texttt{keyboard\_press(key)} & \texttt{key} & element id, key name \\
\texttt{hover}, \texttt{focus}, \texttt{clear}, \texttt{scroll} & unchanged & element id \\
\texttt{goto(url)}, \texttt{go\_back()}, \texttt{go\_forward()} & \texttt{navigate} & url \\
\texttt{drag\_and\_drop(from,\,to)} & \texttt{drag} & both element ids \\
\texttt{upload\_file(bid,\,path)} & \texttt{upload} & element id, path \\
\texttt{send\_msg\_to\_user(m)}, \texttt{report\_infeasible(m)} & \texttt{stop} & message text \\
\texttt{noop()} & \texttt{noop} & --- \\
\midrule
\multicolumn{3}{@{}l}{\emph{\textbf{Role}, from the accessibility tree of the addressed element}} \\
\texttt{link} & \texttt{link} & element name \\
\texttt{button} & \texttt{button} & element name \\
\texttt{checkbox}, \texttt{radio}, \texttt{tab} & unchanged & element name \\
\texttt{option}, \texttt{menuitem}, \texttt{menuitemcheckbox}, \texttt{menuitemradio} & \texttt{option} & element name \\
\texttt{textbox}, \texttt{searchbox} & \texttt{text\_input} & element name \\
\texttt{combobox}, \texttt{listbox} & \texttt{select} & element name \\
any other role & the role verbatim & element name \\
\bottomrule
\end{tabular}
\vspace{-5pt}
\end{table}

\begin{table}[t]
\centering
\small
\setlength{\tabcolsep}{4pt}
\renewcommand{\arraystretch}{0.95}
\caption{\textbf{Canonicalization on MiniWoB++.} The verb is from the agent's primitive action. The role is from the addressed DOM element by the ordered rule below, first matching row wins. The element label (nearest label, ARIA label, text, placeholder, value or id) is kept in the slots. In our corpus the environment reports every input as a typed tag, so the last rule assigns its role.}
\label{tab:canon-miniwob-mind2web}
\begin{tabular}{@{}>{\raggedright\arraybackslash}p{6.2cm}>{\raggedright\arraybackslash}p{3.6cm}>{\raggedright\arraybackslash}p{3.6cm}@{}}
\toprule
\textbf{Raw action or element} & \textbf{Canonical} & \textbf{Kept in slots} \\
\midrule
\multicolumn{3}{@{}l}{\emph{\textbf{Verb}}} \\
\texttt{click(ref)}, \texttt{submit(ref)}, \texttt{clear(ref)} & \texttt{click}, \texttt{submit}, \texttt{clear} & element label \\
\texttt{noop} & \texttt{noop} & --- \\
\texttt{type(ref,\,text)} & \texttt{type} & typed text, element label \\
\texttt{select(ref,\,value)} & \texttt{select} & option value, element label \\
\texttt{key(k)} & \texttt{key} & key name \\
\midrule
\multicolumn{3}{@{}l}{\emph{\textbf{Role}, from the DOM element (first matching rule)}} \\
ARIA role \texttt{button}, \texttt{link}, \texttt{checkbox}, \texttt{radio}, \texttt{tab}, \texttt{dialog} or \texttt{option} & the ARIA role & element label \\
tag \texttt{button}, or \texttt{input} of type \texttt{submit} or \texttt{button} & \texttt{button} & element label \\
tag \texttt{a} & \texttt{link} & element label \\
tag \texttt{select} / \texttt{option} & \texttt{select} / \texttt{option} & element label \\
\texttt{input} of type \texttt{checkbox} / \texttt{radio} & \texttt{checkbox} / \texttt{radio} & element label \\
tag \texttt{textarea}, or \texttt{input} of any other type & \texttt{text\_input} & element label \\
class contains \texttt{tab} / \texttt{dialog}, or tag \texttt{dialog} & \texttt{tab} / \texttt{dialog} & element label \\
tag \texttt{p}, \texttt{span}, \texttt{div}, \texttt{label}, \texttt{td}, \texttt{th}, \texttt{li} & \texttt{text} & element label \\
any other tag, \textit{e.g.,} \texttt{input\_text}, \texttt{input\_checkbox} & the tag name (\texttt{element} if none) & element label \\
\bottomrule
\end{tabular}
\vspace{-10pt}
\end{table}

\begin{table}[t]
\centering
\small
\setlength{\tabcolsep}{3pt}
\renewcommand{\arraystretch}{0.95}
\caption{\textbf{Canonicalization on ScienceWorld and WebShop.} Both
environments take text commands, so the rule is a template match rather than a
role lookup. $X$ and $Y$ are object phrases, and $c(\cdot)$ is the nine-class
lexicon defined in the text.}
\label{tab:canon-sciworld-webshop}
\begin{tabular}{@{}>{\raggedright\arraybackslash}p{5.2cm}>{\raggedright\arraybackslash}p{5.5cm}>{\raggedright\arraybackslash}p{2.6cm}@{}}
\toprule
\textbf{Command} & \textbf{Canonical primitive} & \textbf{Kept in slots} \\
\midrule
\multicolumn{3}{@{}l}{\emph{\textbf{ScienceWorld}}} \\
\texttt{focus on }$X$ & \ctokx{focus}{$c(X)$} & $X$ \\
\texttt{go to }$X$, \texttt{teleport to }$X$ & \ctokx{navigate}{$X$} if $X$ is a room, else \ctokx{navigate}{$c(X)$} & $X$, teleport flag \\
\texttt{open }$X$, \texttt{close }$X$ & \ctokx{open}{$c(X)$}, \ctokx{close}{$c(X)$} & $X$ \\
\texttt{pick up }$X$ & \ctokx{take}{$c(X)$} & $X$ \\
\texttt{put down }$X$, \texttt{drop }$X$ & \ctok{put}{ground} & $X$, surface form \\
\texttt{move }$X$\texttt{ to }$Y$ & \ctokx{put}{$c(Y)$} & $X$, $Y$ \\
\texttt{activate }$X$, \texttt{deactivate }$X$ & \ctokx{activate}{$c(X)$}, \ctokx{deactivate}{$c(X)$} & $X$ \\
\texttt{pour }$X$\texttt{ into }$Y$, \texttt{pour }$X$\texttt{ in }$Y$ & \ctokx{pour}{$c(Y)$} & $X$, $Y$ \\
\texttt{dunk }$X$\texttt{ into }$Y$, \texttt{dunk }$X$\texttt{ in }$Y$ & \ctokx{dunk}{$c(Y)$} & $X$, $Y$ \\
\texttt{mix }$X$ & \ctokx{mix}{$c(X)$} & $X$ \\
\texttt{look around} & \ctok{observe}{room} & --- \\
\texttt{look at }$X$, \texttt{look in }$X$, \texttt{examine }$X$ & \ctokx{observe}{$c(X)$} & $X$, surface form \\
\texttt{inventory} & \ctok{observe}{inventory} & --- \\
\texttt{task}, \texttt{reset task} & \ctok{observe}{task} & surface form \\
\texttt{read }$X$ & \ctokx{read}{$c(X)$} & $X$ \\
\texttt{use }$X$\texttt{ on }$Y$ & \ctokx{measure}{$c(Y)$} if $X$ is a thermometer, stopwatch or ruler, else \ctokx{use}{$c(Y)$} & $X$, $Y$ \\
\texttt{use }$X$ & \ctokx{use}{$c(X)$} & $X$ \\
\texttt{connect }$X$\texttt{ to }$Y$ & \ctokx{connect}{$c(Y)$} & $X$, $Y$ \\
\texttt{disconnect }$X$ & \ctokx{disconnect}{$c(X)$} & $X$ \\
\texttt{eat }$X$, \texttt{flush }$X$ & \ctokx{eat}{$c(X)$}, \ctokx{flush}{$c(X)$} & $X$ \\
\texttt{wait1}, \texttt{wait} & \ctok{wait}{short}, \ctok{wait}{long} & --- \\
a bare integer & \ctok{disambiguate}{choice} & the integer \\
anything else (catch-all) & the first word as verb, the rest as role & rest of the command \\
\midrule
\multicolumn{3}{@{}l}{\emph{\textbf{WebShop}}} \\
\texttt{search[}$q$\texttt{]} & \ctok{search}{query} & query text \\
\texttt{click[}$X$\texttt{]}, $X$ a  product id & \ctok{click}{item} & the product id \\
\texttt{click[}$X$\texttt{]}, $X$ a value the goal names & \ctokx{click}{\texttt{option-}\textit{type}} & the value \\
\texttt{click[}$X$\texttt{]}, any other value & \ctok{click}{option} & the value \\
\texttt{click[buy now]} & \ctok{click}{buy} & --- \\
\texttt{click[back to search]} & \ctok{click}{back} & --- \\
\texttt{click[next\,$>$]}, \texttt{click[$<$\,prev]} & \ctok{click}{next}, \ctok{click}{prev} & --- \\
\texttt{click[description]} & \ctok{click}{desc} & --- \\
\texttt{click[features]} & \ctok{click}{features} & --- \\
\texttt{click[reviews]} & \ctok{click}{reviews} & --- \\
\texttt{click[attributes]} & \ctok{click}{attrs} & --- \\
\bottomrule
\end{tabular}
\vspace{-20pt}
\end{table}

The purpose is to better recognize reusable structures in the corpus. Tables~\ref{tab:canon-webarena}
to~\ref{tab:canon-sciworld-webshop} show the templates from a raw action to a canonical token.
Properties of the template are three-fold: \textbf{1)} \emph{deterministic}: written once per environment from the action space, 
\textbf{2)} \emph{complete}: no action can fail to canonicalize and none is silently dropped, and \textbf{3)} \emph{no loss of information}: every trimmed value, \textit{e.g.,} an element id, typed text, an object name, is kept so it could still be used after \xtree{} mining.
Chain-of-Thought never enters the alphabet.

On ScienceWorld the object class $c(\cdot)$ is an ordered keyword lexicon of
nine classes, tried in order: \texttt{door}, \texttt{instrument},
\texttt{paint}, \texttt{container}, \texttt{device}, \texttt{plant},
\texttt{animal}, \texttt{substance}, \texttt{location}, with the literal head noun kept when nothing matches and \texttt{location} the closed list of the ten ScienceWorld rooms. Instance numbers and containment qualifiers are stripped before the match and preserved. This is what makes \texttt{move
thermometer to metal pot} and \texttt{move thermometer to ceramic cup} the same
symbol, \ctokx{put}{$c(Y)$}$\,={}$\ctok{put}{container}, while keeping both
distinct from \texttt{move thermometer to stove}. Simplified examples in Figure \ref{fig:concept} and \ref{fig:exemplar} are for illustration purpose.

\section{A case study}
\label{app:casestudy}

Tables~\ref{tab:case-webarena}--\ref{tab:case-webshop} study \xtree{} nodes from the three environments to illustrate the semantics.
First, the boundaries are meaningful: the most frequent
WebArena depth-2 node is a two-field form followed by its submit button, and
the same canonical node instantiates as a report filter on the shop admin
site and as a route query on the map, which are reusable structures of real-world meanings. 
Second, depth composes: the ScienceWorld chain S8
$\to$ S19 $\to$ S20 $\to$ S21 adds one primitive per depth, from ``go to the
kitchen via the hallway'' to ``fetch the thermometer,'' so a depth-6 node is a sub-procedure while its descendants remain reusable on their own. 
Tree-structured decomposition also helps answer complex questions over knowledge bases \citep{huang2023qdt}. \xtree{} mines such a tree from trajectories instead.
Third, the \xscore{} separates structures that the
verifier rewards from those it does not: on WebShop, the node that engage an option type named by the goal (color, size, flavor) succeed in 81--97\% of their occurrences, the universal macros (search, click item, buy) sit at the corpus mean of 0.56, and the pagination chains and untyped option clicks sit
at 0.14--0.36. This is the separation the \xscore{} merge and the
option-recall gate exploit (\S\ref{sec:webshop-signal}). The libraries also
contain low-value macros (WebArena's menu-then-submenu click pair recurs
1{,}219 times), which is why the offline reward scales with depth.

\begin{figure}[t]
\centering
\includegraphics[width=\textwidth]{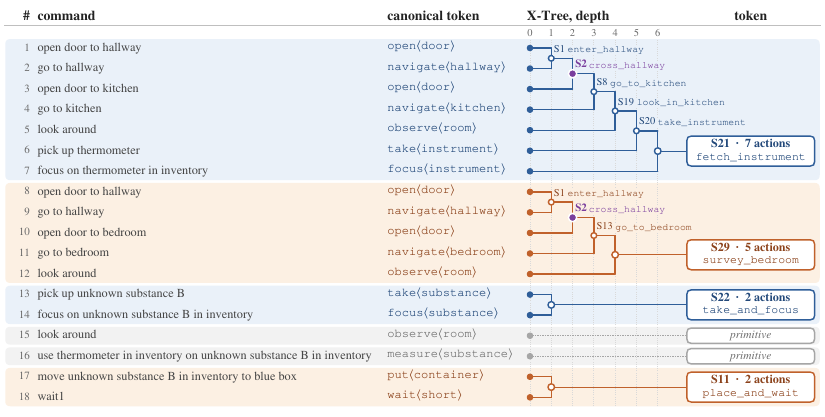}
\vspace{-20pt}
\caption{\textbf{An example trajectory ``tokenized'' by the \xtree{} (ScienceWorld).} Task: \textit{measure the temperature of unknown substance B in the bedroom, then place it in the green box above 50 degrees and in the blue box below.} Rows
are the trajectory's 18 actions with their canonical tokens. The tree beside them is the \xtree{} over the actions,
with depth growing to the right, and each shaded band is one token of the tiling. S2 (purple) recurs inside two
different tokens.}
\label{fig:trajectory-tree}
\vspace{-6pt}
\end{figure}

\begin{table}[t]
\centering
\scriptsize
\setlength{\tabcolsep}{2pt}
\renewcommand{\arraystretch}{1.08}
\vspace{8pt}
\caption{\textbf{Case study (a): \xtree{} nodes in WebArena.} Canonical nodes are written in the \ctok{verb}{role}
tokens of Table~\ref{tab:canon-webarena}. Occ.\ counts the occurrence of the node. The
grounded occurrence reads element labels from the page's accessibility tree. }
\label{tab:case-webarena}
\begin{tabular}{@{}>{\raggedright\arraybackslash}p{3.6cm}cc>{\raggedright\arraybackslash}p{5.8cm}>{\raggedright\arraybackslash}p{3.cm}@{}}
\toprule
\textbf{Canonical node} & \textbf{Depth} & \textbf{Occ.} & \textbf{Grounded occurrence} & \textbf{Reading} \\
\midrule
\ctok{type}{text\_input} $\to$ \ctok{type}{text\_input} $\to$ \ctok{click}{button} & 2 & 209 & fill ``From *'' 01/01/2025; fill ``To *'' 03/24/2025; click ``Show Report'' (shop admin); fill ``From'' Times Square; fill ``To'' Central Park; click ``Go'' (map) & complete a two-field form and submit: a report filter and a route query are the same node \\
\ctok{type}{text\_input} $\to$ \ctok{click}{button} & 1 & 430 & fill searchbox ``Search for projects, issues, etc.'' \emph{project}; click ``Search'' (GitLab) & type a query and submit it \\
\ctok{type}{text\_input} $\to$ \ctok{type}{text\_input} & 1 & 88 & fill ``From'' Grand Central Station; fill ``To'' Empire State Building (map) & the two-field entry that recurs inside the first row \\
\ctok{click}{link} $\to$ \ctok{click}{link} $\to$ \ctok{click}{link} & 2 & 161 & click ``Forums''; click a page link; click the next page (Reddit) & navigate, then paginate \\
\ctok{click}{link} $\to$ \ctok{click}{link} & 1 & 1{,}219 & click ``SALES''; click a submenu entry (shop admin) & menu navigation \\
\bottomrule
\end{tabular}
\vspace{-9pt}
\end{table}

\begin{table}[t]
\centering
\scriptsize
\setlength{\tabcolsep}{2pt}
\renewcommand{\arraystretch}{1.08}
\caption{\textbf{Case study (b): \xtree{} nodes in ScienceWorld.} ScienceWorld nodes are long, so each canonical
node is written as its merge of two children, in the primitives of Table~\ref{tab:canon-sciworld-webshop}. Occ.\ counts
occurrences in the $1{,}673$ AgentTraj-L trajectories.}
\label{tab:case-sciworld}
\begin{tabular}{@{}>{\raggedright\arraybackslash}p{3.6cm}cc>{\raggedright\arraybackslash}p{5.0cm}>{\raggedright\arraybackslash}p{3.5cm}@{}}
\toprule
\textbf{Canonical node} & \textbf{Depth} & \textbf{Occ.} & \textbf{Grounded occurrence} & \textbf{Reading} \\
\midrule
S8$=$S2$+$\allowbreak \ctok{navigate}{kitchen} & 3 & 415 & open door to hallway; go to hallway; open door to kitchen; go to kitchen & go to the kitchen via the hallway \\
S31$=$\ctok{put}{device}$+$\allowbreak S30 & 4 & 151 & move metal pot containing gallium to stove; activate stove; examine gallium; use thermometer on gallium; examine; use thermometer & heat a substance and monitor its temperature \\
S60$=$\ctok{navigate}{foundry}$+$\allowbreak S59 & 6 & 25 & go to foundry; open blast furnace; move metal pot to blast furnace; activate blast furnace; examine lead; use thermometer on lead; examine; use thermometer & melt a metal and monitor it \\
S5$=$\ctok{observe}{room}$+$\allowbreak S4 & 4 & 381 & look around; connect battery anode to black wire terminal 1; connect battery cathode to blue wire terminal 1; connect black wire terminal 2 to cathode in buzzer; \ldots & wire a circuit from battery through wires to a load \\
S27$=$S26$+$\allowbreak \ctok{take}{animal} & 4 & 241 & open door to outside; go to outside; look around; focus on baby wolf; pick up baby wolf & find and collect an animal outdoors \\
S43$=$S7$+$\allowbreak \ctok{put}{container} & 4 & 90 & open door to hallway; go to hallway; open door to workshop; go to workshop; move baby wolf in inventory to blue box & carry the animal to a container in another room \\
S15$=$\ctok{wait}{short}$+$\allowbreak S14 & 3 & 381 & wait1; look around; move aluminum foil to blue box; wait1 & observe, then sort an object into a box \\
\bottomrule
\end{tabular}
\vspace{-15pt}
\end{table}

\begin{table}[t]
\centering
\scriptsize
\setlength{\tabcolsep}{2pt}
\renewcommand{\arraystretch}{1.08}
\vspace{5pt}
\caption{\textbf{Case study (c): \xtree{} nodes in WebShop.} Canonical nodes are written in the primitives of
Table~\ref{tab:canon-sciworld-webshop}. Occ.\ counts occurrences in the $1{,}824$ ETO trajectories. ETO is the corpus with failures, so the fraction of a node's occurrences in successful trajectories (corpus mean $0.56$) is reported.}
\label{tab:case-webshop}
\begin{tabular}{@{}>{\raggedright\arraybackslash}p{6.8cm}ccc>{\raggedright\arraybackslash}p{4.4cm}@{}}
\toprule
\textbf{Canonical node} & \textbf{Depth} & \textbf{Occ.} & \textbf{Success} & \textbf{Reading} \\
\midrule
\ctok{search}{query} $\to$ \ctok{click}{item} $\to$ \ctok{click}{option-color} $\to$ \ctok{click}{buy} & 3 & 94 & 0.81 & engage the goal's color option before buying \\
\ctok{search}{query} $\to$ \ctok{click}{item} $\to$ \ctok{click}{option-size} $\to$ \ctok{click}{option-color} $\to$ \ctok{click}{buy} & 4 & 25 & 0.92 & two typed options, both named by the goal \\
\ctok{search}{query} $\to$ \ctok{click}{item} $\to$ \ctok{click}{option-flavor-name} & 2 & 31 & 0.97 & a typed option the goal names \\
\ctok{click}{option-color} $\to$ \ctok{click}{buy} & 1 & 139 & 0.84 & the closing move once the right option is set \\
\ctok{search}{query} $\to$ \ctok{click}{item} $\to$ \ctok{click}{buy} & 2 & 741 & 0.57 & buy without touching options: a universal macro at the corpus mean \\
\ctok{search}{query} $\to$ \ctok{click}{item} & 1 & 1{,}770 & 0.57 & a universal macro \\
\ctok{search}{query} $\to$ \ctok{click}{next} $\to$ \ctok{click}{next} $\to$ \ctok{click}{next} $\to$ \ctok{click}{item} & 4 & 10 & 0.30 & paging deep into the results: failure-associated \\
\ctok{search}{query} $\to$ \ctok{click}{next} $\to$ \ctok{click}{item} $\to$ \ctok{click}{buy} & 3 & 14 & 0.14 & buying after paging past the first results \\
\ctok{search}{query} $\to$ \ctok{click}{item} $\to$ \ctok{click}{option} $\to$ \ctok{click}{option} $\to$ \ctok{click}{option} $\to$ \ctok{click}{buy} & 4 & 14 & 0.36 & untyped options clicked in sequence \\
\bottomrule
\end{tabular}
\vspace{-8pt}
\end{table}

\paragraph{One trajectory through the tree.}
\label{app:example}
Figure~\ref{fig:trajectory-tree} tokenizes one trajectory with the ScienceWorld \xtree{}.
The task is \textit{to measure the temperature of an unknown substance and put it in a box that depends on the reading}.
The agent starts in the art studio and fetches the thermometer from the kitchen, which reads 1\,$^\circ$C, below the task's threshold of 50 degrees, so it uses the blue box.
That last choice rests on a numerical comparison.
The merges turn the 18 actions into six tokens: four \xtree{} nodes and two primitives that stay unmerged.
The first token, S21, is the depth-$6$ chain of Table~\ref{tab:case-sciworld} that fetches the thermometer from the kitchen.
The second, S29, walks to the bedroom and looks around.
Both are built on S2 (open the door to the hallway, go there, open the next door), which occurs $1{,}803$ times in the corpus and here completes once toward the kitchen and once toward the bedroom.
Under the node-per-row SFT format of Appendix~\ref{app:sft}, each token is one training row, so this trajectory gives six rows instead of $18$.

\section{\xtree{} as skills in the context}
\label{app:prompted}
\begin{wraptable}{r}{0.52\textwidth}
\vspace{-22pt}
\centering
\footnotesize
\setlength{\tabcolsep}{4pt}
\caption{\textbf{\xtree{} as advice in context} (MiniWoB++, SR, \%). Only the skill bank in context changes. Four LLM-written baselines and \xtree{}.}
\label{tab:prompted}
\begin{tabular}{@{}lccc@{}}
\toprule
\cmidrule(l){2-4}
\textbf{Skill bank} & \textbf{Qwen2.5-7B} & \textbf{Qwen3-8B} & \textbf{Qwen3-14B} \\
\midrule
no skills   & 44.9 & 60.4 & 68.1 \\
AWM         & 41.4 & 59.9 & 69.0 \\
SkillWeaver & 47.6 & 61.3 & 68.3 \\
WALT        & 46.3 & 58.1 & 68.5 \\
WebXSkill   & 43.8 & 59.8 & 68.6 \\
\xtree{}    & \textbf{49.8} & \textbf{65.1} & \textbf{72.6} \\
\bottomrule
\end{tabular}
\vspace{-10pt}
\end{wraptable}

We test \xtree{} as as advice in context under an interaction harness \citep{huang2024queryagent,cheng2024readi} shared with four
LLM-abstraction baselines (AWM \citep{wang2024awm},
SkillWeaver \citep{zheng2025skillweaver}, WALT \citep{prabhu2026walt}, WebXSkill \citep{wang2026webxskill}) and a no-skill control.
Only the induction algorithm differs. Other settings are identical. We re-implement the baselines based on the same harness and base model for fair comparison. We use the same model as inference to mine skills for the baselines. Note that our \xtree{} requires zero LLM calls. Table~\ref{tab:hyper-prompted} lists the detailed settings. Results are shown in Table \ref{tab:prompted}.

On MiniWoB++, \xtree{} leads every baseline at three backbones, by
up to $3.8\%$, while some baselines fall below the no-skill setting, since the LLMs that write the skills are relatively small. 
Agents tend to conform to misleading input \citep{zhou2026epistemic}. And how a model combines the context with its own knowledge is uncertain \citep{cheng2024interplay,jin2024disentangling}. Training the skills into the weights, as in our main experiments, avoids that dependence.
This validates the quality of reusable structure in \xtree{} without integration with training, built deterministically without LLM calls.

\section{\xtree{} integration with SFT}
\label{app:sft}
\paragraph{Segmentation.}
Standard SFT trains one action per training row with a full context (system prompt, goal, observation, history) for every row.
We test \xtree{} integration with SFT by \emph{one \xtree{} node per row}: the target is the node's full action span (several actions generated one time) and the context is encoded once per node. A length-matched sliding window baseline is the random cut over action stream with matched length as \xtree{} node. The three formats supervise the identical action streams (\textit{i.e.,} trained over the same trajectory corpus). On the ScienceWorld 300-random-episode corpus, the vanilla SFT costs 1{,}766 tokens per supervised action against 446 for the \xtree{} node SFT, a $4.0\times$ saving, with $97.6\%$ of raw's tokens being repeated prompts under the same training settings.

\paragraph{Test-time protocol.}
A node-trained policy outputs one turn per node: a \texttt{<planning>} block and one \texttt{<action>} tag per action. The evaluator executes the tags in order, one environment step each, and returns no observation until the turn ends (the last action is executed). The next prompt carries the observation after the last executed action and the turn's joined action string, which is the same node-level history the SFT rows are built with. An action that the environment rejects returns its error message as that step's observation. The remaining tags still execute. Then, the policy sees the failure at its next turn. Every run has the same budget of 30 environment actions per episode. The standard SFT row executes one action per turn as in \citep{zhang2025rlvmr}.

\paragraph{SFT integration with \xtree{} outperforms the standard one at all scales (Table~\ref{tab:sciworld-sft}).}\begin{wraptable}{r}{0.42\textwidth}
\vspace{-10pt}
\centering
\footnotesize
\setlength{\tabcolsep}{3pt}
\caption{\textbf{SFT at 1.5B/3B/7B on ScienceWorld} (SR pooled over G0 and G1). 
}
\label{tab:sciworld-sft}
\begin{tabular}{@{}lccc@{}}
\toprule
\textbf{SFT row} & \textbf{1.5B} & \textbf{3B} & \textbf{7B} \\
\midrule
one standard action  & 11.7 & 16.7 & 20.8 \\
length-matched span & 12.0 & 17.4 & 19.3 \\
an {\xtree{} node} & \textbf{16.6} & \textbf{19.1} & \textbf{29.9} \\
\bottomrule
\end{tabular}
\vspace{-8pt}
\end{wraptable}

While consuming $4\times$ fewer tokens for the identical supervised actions, our \xtree{} with SFT leads by $+4.9$, $+2.4$ and $+9.1$.
The length-matched baseline further validates the effectiveness of \xtree{}: a sliding window with the same span lengths is on par with the vanilla SFT, under-performing \xtree{} significantly. This indicates that \xtree{} truly merges meaningful reusable structures that guides training to save compute budget and improve performance.

\section{Offline RL on WebArena}
\label{app:webarena}

\subsection{Rollout observations for offline RL}\label{app:rollout_details}
We have the trajectory pool providing the thought, action and observation at each interaction turn. At an \xtree{} node boundary the policy is
given the gold prefix (\textit{i.e.,} previous observation and action steps) and output one action per turn.
If the action reproduces the exact recorded gold action (thought excluded) and the node has steps left, the recorded next observation in the trajectory is appended and the rollout continues. If it does
not, the rollout ends there. 
So the loop never has to invent an observation for
a state the gold trajectory does not contain. 
The step-matching reward of Eq.~\ref{eq:offline-reward} is the fraction of the node's steps reached before
that stop. Observations are replayed in Go-Browse's compact form and the whole
prefix is retained.

\subsection{Evaluation protocol}
\begin{table}[t]
\centering
\vspace{-5pt}
\caption{Official evaluator and normalized SR (WebArena, Qwen2.5-7B, mean over three passes).}
\label{tab:webarena-sr}
\small
\setlength{\tabcolsep}{4pt}
\begin{tabular}{@{}lcc@{}}
\toprule
 \textbf{Method} & \textbf{official SR} & \textbf{SR (main text)} \\
\midrule
 base model & 5.8 & 6.5 \\
  SFT-Full (Go-Browse), 2 epochs & 17.2 & 18.4 \\
  SFT-Full, 4 epochs & 17.4 & 18.8 \\
  whole-trajectory GRPO & 18.7 & 19.9 \\
  whole-trajectory GRPO, binary outcome reward & 17.6 & 18.2 \\
  random-node GRPO & 18.5 & 19.5 \\
  random-tree GRPO & 17.0 & 17.9 \\
  \xtree{} plain mixing & 19.3 & 20.7 \\
  \xtree{} sequential & 20.2 & 21.2 \\
  \xtree{} mixing + hier.\ reward & 21.5 & 22.9 \\
\bottomrule
\end{tabular}
\vspace{-10pt}
\end{table}

Every run is scored on the same 694 deterministic tasks (the fuzzy-match tasks of the 812-task set are excluded) under one harness, three passes per run with greedy decoding. The evaluation has no randomness. The pass-to-pass variance comes from environment state and service timing. All sites are reset between runs. We report normalized SR mean over passes. Every
per-site row is validated against its pass totals. Statistical significance is in Appendix \ref{app:stability}.
Table~\ref{tab:webarena-sr} lists the
official evaluator's success rate next to the normalized SR (Qwen2.5-7B). The normalizer is frozen in code and applied identically to all runs. It recovers 4--11 of the 45 \texttt{exact\_match} tasks per pass. 

The WebArena \xtree{} node set used for training is $5{,}148$ \xtree{} nodes, $3.5$k$/1.2$k$/0.4$k of them at depth $1$/$2$/$3$ (the three levels take most of the nodes). The tree-block of
Table~\ref{tab:webarena-ablations} are built from that same node source and
trained from the same SFT warm start with the settings of
Table~\ref{tab:hyper-rl}. Random spans match the \xtree{} nodes exactly in row count ($5{,}148$) and in horizon histogram. The
random tree matches the row count but is shallower by construction since it replaces \xscore{} by random pairing (88\% of its nodes are two actions, against 43\% for the mined tree). The whole-trajectory
baseline has one span per trajectory ($1{,}989$ rows), so we train for equivalent gradient steps to match the compute with our recipe for fair comparison.

The released Go-Browse-WA-7B model scores 18.0 SR (19.0 normalized SR) on our 694
deterministic tasks as the mean of three passes, against a published 21.7\% on
the 812-task set under the same harness. The gap is within what the
excluded fuzzy-match tasks and the harness differences account for. Every
run in the paper is scored under the same harness.

\subsection{Transferability to larger scale and model family}
\begin{wraptable}{r}{0.47\textwidth}
\vspace{-23pt}
\centering
\footnotesize
\setlength{\tabcolsep}{3.8pt}
\caption{{Scale and model-family transfer of offline RL with \xtree{}} (WebArena, SR, \%).}
\label{tab:scale-family}
\begin{tabular}{@{}lccc@{}}
\toprule
\textbf{Backbone} & \textbf{SFT-Full} & \textbf{$+$\xtree{} recipe} & $\Delta$ \\
\midrule
Qwen2.5-7B & 18.4 & \textbf{22.9} & $+4.5$ \\
Qwen2.5-14B & 24.1 & \textbf{25.7} & $+1.6$ \\
GLM-4-9B & 14.9 & \textbf{16.9} & $+2.0$ \\
\bottomrule
\end{tabular}
\vspace{-8pt}
\end{wraptable}

Table~\ref{tab:scale-family} shows the transferability of our recipe
 discussed in \S\ref{sec:webarena}. On Qwen2.5-14B, our recipe adds
$+1.6$ over full-data SFT. On GLM-4-9B, it adds $+2.0$ over the SFT-full recipe. This demonstrates that our offline RL integration with \xtree{} can transfer to a larger scale and another model family, further showing the effectiveness of \xtree{} and the integration method.

\section{Online RLVR on ScienceWorld}
\label{app:sciworld}

\subsection{Quantitative analysis of the trained policy on \xtree{} skills}
\begin{figure}[t]
\centering
\begin{minipage}[c]{0.5\textwidth}
\centering
\includegraphics[width=\linewidth]{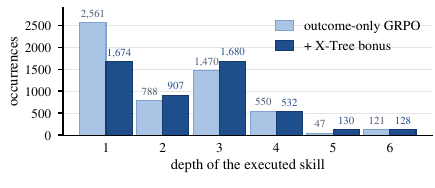}
\end{minipage}\hfill
\begin{minipage}[c]{0.47\textwidth}
\centering\footnotesize
\setlength{\tabcolsep}{3pt}
\resizebox{\linewidth}{!}{%
\begin{tabular}{@{}lcc@{}}
\toprule
 & \textbf{outcome-only} & \textbf{$+$\xtree{}} \\
\midrule
episodes with a mined skill & 94.6\% & 99.3\% \\
distinct skills executed & 51 & 50 \\
skills per episode & 3.33 & 3.04 \\
mean executed depth & 2.11 & \textbf{2.39} \\
episodes with a depth$\ge$3 skill & 85.4\% & \textbf{87.5}\% \\
\bottomrule
\end{tabular}}
\end{minipage}
\vspace{-10pt}
\caption{{What the trained policies execute at test time} (1.5B, G0 fold). Left: occurrences of executed \xtree{} skills by depth.
Right: properties of the trained models.}
\vspace{-5pt}
\label{fig:executed-depth}
\end{figure}

We test whether the \xtree{} skills end up in the trained policy. We trace the actions the trained policy takes at evaluation and match them against the \xtree{} nodes the reward matched (1.5B on G0 Figure~\ref{fig:executed-depth}).
We find that \textit{skill coverage does not separate the two policies, but the depth does.} The outcome-only policy executes a comparable number of \xtree{} skills in its episodes. However, the \xtree{} bonus policy's histogram moves mass out of lower depth into depth-$3$ and depth-$5$. The mean executed depth rises from $2.11$ to $2.39$, and our recipe executes \emph{fewer} skills per episode, which indicates that the policy runs fewer and longer composites. This matches the finding that RL composes the atomic skills a model already holds into larger ones \citep{cheng2025atomic}.

\subsection{The dynamic of $\lambda_g$ over training steps}
\label{app:rewards}
\begin{figure}[t]
\centering
\includegraphics[width=\textwidth]{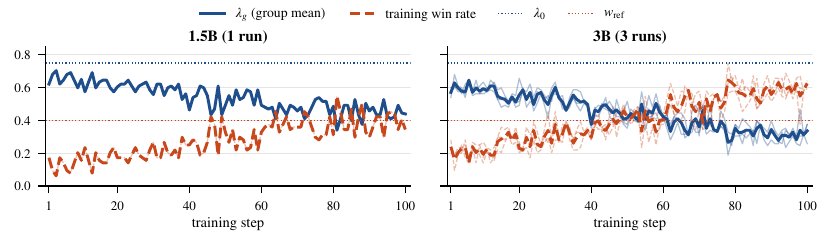}
\vspace{-22pt}
\caption{\textbf{How $\lambda_g$ anneals during training (ScienceWorld).} Per training
step, the group-mean weight $\lambda_g$ of Eq.~\ref{eq:online-reward} (solid) and the
training win rate (dashed). $\lambda_0{=}0.75$ and $w_{\mathrm{ref}}{=}0.4$ are dotted.
3B: the three runs behind Figure~\ref{fig:online-rlvr} (faint) and their mean (bold).
1.5B: one run.}
\vspace{-10pt}
\label{fig:anneal}
\end{figure}

The weight $\lambda_g$ in Eq.~\ref{eq:online-reward} is a dynamic property of training rather than a direct add-on. It down-weights the bonus as a task's group win rate approaches $w_{\mathrm{ref}}$. Figure~\ref{fig:anneal} traces the logged group-mean $\lambda_g$ against the training win rate on ScienceWorld, step by step, for runs of Figure~\ref{fig:online-rlvr}. The two scales anneal at the rates the learning curves dictate:
the 3B win rate rises from $\sim0.2$ to $\sim0.6$ and the mean $\lambda_g$ falls from $0.57$ to $0.34$, while at 1.5B the win rate ends at
$0.34$ and $\lambda_g$ is $0.44$. On WebShop, where $w_{\mathrm{ref}}$
is $0.85$, the dynamic takes $\lambda_g$ from $0.71$ at a $2\%$ win rate to
$0.11$ at $72\%$. $\lambda_g$ never reaches zero on a task the policy has not learned perfectly, which is within expectation: the shaping fades by competence, not by step count.
Reward structures can also be discovered by automated search \citep{cheng2025derl}. Ours keeps a fixed form and adapts only its weight to the win rate.

\subsection{Sensitivity to $\lambda_0$ and $w_{\mathrm{ref}}$}
\label{app:lambda_sens}
\begin{wraptable}{r}{0.45\textwidth}
\vspace{-22pt}
\centering
\footnotesize
\setlength{\tabcolsep}{4pt}
\caption{\textbf{Sensitivity to $\lambda_0$ and $w_{\mathrm{ref}}$} (ScienceWorld, 1.5B, $n{=}8$, SR on G0 / G1).}
\label{tab:lambda-sens}
\resizebox{\linewidth}{!}{%
\begin{tabular}{@{}lccc@{}}
\toprule
\textbf{Setting} & $\lambda_0$ & $w_{\mathrm{ref}}$ & \textbf{G0 / G1} \\
\midrule
outcome-only GRPO & -- & -- & 38.6 / 36.2 \\
gentler weight & 0.5 & 0.4 & 39.4 / 37.6 \\
front-loaded, anneals sooner  & 1.5 & 0.25 & 38.9 / 36.4 \\
selected (Ours) & 0.75 & 0.4 & {39.8} / {39.5} \\
\bottomrule
\end{tabular}}
\vspace{-8pt}
\end{wraptable}

The two hyperparameters of Eq.~\ref{eq:online-reward} are selected by the
training-side validation: take the run with the highest validation success points. No testing set numbers enter the choice. We ran two ablations in opposite directions at
1.5B, $n{=}8$ (Table~\ref{tab:lambda-sens}): a gentler weight and a
front-loaded one that weights early episodes more and anneals sooner. Both
remain comparable to or above outcome-only GRPO and below our setting. So the recipe is not very sensitive to the choice of both hyperparameters.

\subsection{Ablation of \xtree{} structure}
We ablate \xtree{} by replacing the skill set the adaptive bonus reads
(Table~\ref{tab:random-library}). The first ablation draws spans at arbitrary positions, which leaves the bonus far fewer matching. The second control keeps everything except \xscore{} to merge: $80$ skills with the length and depth
histograms of \xtree{}, each drawn at random from canonical action pairs. Other settings are fixed for fair comparison.

\begin{wraptable}{r}{0.47\textwidth}
\vspace{-22pt}
\centering
\footnotesize
\setlength{\tabcolsep}{3pt}
\caption{Ablation of \xtree{} (ScienceWorld, 1.5B). SR (\%), difference to outcome-only in parentheses. ``\#Match'' is the number of nodes the bonus can match.}
\label{tab:random-library}
\resizebox{\linewidth}{!}{%
\begin{tabular}{@{}lrccc@{}}
\toprule
\textbf{Skills} & \textbf{\# Match} & \textbf{G0} & \textbf{G1} & \textbf{G2} \\
\midrule
Empty (outcome-only) & -- & 38.6 & 36.2 & 22.8 \\
Random spans  & 576 & 38.4 \scriptsize$(-0.2)$ & 38.4 \scriptsize$(+2.2)$ & -- \\
Random tree& 2{,}109 & 38.1 \scriptsize$(-0.5)$ & 36.8 \scriptsize$(+0.6)$ & \textbf{26.7} \scriptsize$(+3.9)$ \\
\xtree{} & 2{,}698 & \textbf{39.8} \scriptsize$(+1.2)$ & \textbf{39.5} \scriptsize$(+3.3)$ & 25.6 \scriptsize$(+2.8)$ \\
\bottomrule
\end{tabular}}
\vspace{-8pt}
\end{wraptable}

The ablation validates our \xtree{} structure. On G0, only our \xscore{} mined bank is above outcome-only GRPO ($+1.2$ against $-0.5$ and $-0.2$ for the two controls). On G1, our \xtree{} leads by
$+3.3$, with random spans recovering $+2.2$ of it, indicating the superiority of \xtree{}. On the held-out fold, three banks are comparable. We therefore
attribute the bonus's gain to the \xtree{} structure mined with \xscore{}.

\subsection{Robustness of \xtree{} from a held-out mining corpus}
\label{app:leakage}
\begin{table}[t]
\centering
\footnotesize
\caption{\textbf{Results with held-out tasks removed from the mining corpus} (ScienceWorld,
1.5B). Top: \xtree{} changes from the held-out tasks removed. Bottom: performance with \xtree{} original and re-mined, everything else fixed.}
\label{tab:leakage}
\small
\setlength{\tabcolsep}{5pt}
\begin{tabular}{@{}lcc@{}}
\toprule
 & \textbf{full corpus} & \textbf{testing task types removed} \\
\midrule
trajectories after the length filter & 1{,}673 & 1{,}248 \\
\# \xtree{} nodes & 80 & 80 (66 kept, 14 new) \\
grounded skills the bonus reward can match & 2{,}698 & 1{,}305 \\
\midrule
G2 SR, outcome $+$ \xtree{} bonus (three runs) & \textbf{25.6} & 24.8 \\
G2 SR, outcome-only GRPO & \multicolumn{2}{c}{22.8} \\
\bottomrule
\end{tabular}
\vspace{-10pt}
\end{table}

In the main part, we test \xtree{} on ScienceWorld with three generalization levels \citep{zhang2025rlvmr,huang2024targa,huang2023markqa}. A question is whether \xtree{} mined from a corpus never covering the held-out task can transfer. We test by removing the testing types from the mining corpus (Table~\ref{tab:leakage}).
We observe that the grounded skills our bonus reward matches drops by about half.
Training the held-out runs (G2, 1.5B) with the re-mined \xtree{} with other settings fixed, \xtree{} still lead over outcome-only GRPO by $+2.0$, showing that \xtree{} still works well with a smaller matching surface. This suggests the robustness and transferability of \xtree{}, and that \xtree{} does not require a strong and large scale trajectory pool to perform, echoing findings in Table \ref{tab:webshop-variants}.

\section{OPSD on WebShop}
\label{app:opsd}

\subsection{The \xtree{} rendered privileged context}
\label{app:render}
\begin{table}[t]
\centering
\caption{\textbf{The \xtree{} rendering example for the color option}, as the self-teacher reads it.}
\label{tab:render-color}
\setlength{\tabcolsep}{6pt}
\renewcommand{\arraystretch}{1.0}
\begin{tabular}{p{\dimexpr\linewidth-2\tabcolsep-2\arrayrulewidth\relax}}
\hline
\rule{0pt}{2.2ex}\ttfamily\fontsize{7.5}{8.9}\selectfont\raggedright
\#\#\# TASK: goal requires a color option \#\#\#\\
The instruction names a color. Expert episodes that selected the goal's color on the product page before buying, and how they fared:\\
- search<query> -> click<item> -> click<option-color> -> click<buy>: 81\% success (94 episodes, ABOVE base) -- search for the product; open a result item; select the goal's color option; click Buy Now.\\
- click<option-color> -> click<option> -> click<buy>: 62\% success (34 episodes, ABOVE base) -- select the goal's color option; select an option the goal does NOT ask for (wasted step); click Buy Now.\\
\textrm{\itshape [four more chains]}\\
Plan: search with the product words plus the color value; open a matching item; click the exact color value the instruction asks for (it must be visible on the page); select any other required options the same way; then Buy Now.\vspace{3pt}
\tabularnewline \hline
\end{tabular}
\vspace{-15pt}
\end{table}

The self-teacher reads \xtree{} as plain text by a fixed rendering template with no LLM calls. The template produces one rendering per goal-option type (color, size, flavor, etc) and one general rendering. A rendering opens with a header naming the option type, then lists up to six mined \xtree{}-node-based chains selecting that option, ranked by corpus success rate, one line each in a fixed format: the canonical chain, success rate and occurrence count, a tag against the corpus base rate (\texttt{above}, \texttt{at}, \texttt{below}), and a gloss that maps every token to a fixed phrase (\ctok{click}{option-color} to
``select the goal's color option'', and a \ctok{click}{option} the goal did not ask
for to ``wasted step''). It closes with a one-sentence plan, which states the task's rules outright \citep{zhou2025rulearena}. Table~\ref{tab:render-color} shows the rendering for the color option.

The general rendering states the corpus and lists the option-free chains in the same
format. Retrieval is the released trainer's keyword rule: an instruction that
contains ``color:'' receives the general rendering followed by the color
rendering, prepended to teacher's prompt.
Organizing procedural knowledge into such logic units helps the models understanding \citep{an2025thread}, and a better retriever is worth future exploration \citep{zhuang2024efficientrag}. 

\subsection{Ablation on skill bank to the self-teacher}
\begin{wraptable}{r}{0.36\textwidth}
\vspace{-22pt}
\centering
\footnotesize
\setlength{\tabcolsep}{5pt}
\caption{\textbf{Ablation on skill bank to the self-teacher (WebShop, 3B).}}
\label{tab:opsd-bank-controls}
\begin{tabular}{@{}lc@{}}
\toprule
\textbf{Teacher's bank} & \textbf{Success} \\
\midrule
\multicolumn{2}{@{}l}{\emph{outcome-only RLVR: $74.0$}} \\
\midrule
Empty (no privileged text) & 74.2 \\
\xtree{}, words scrambled & 76.5 \\
Mined \xtree{} & \textbf{77.1} \\
\bottomrule
\end{tabular}
\vspace{-10pt}
\end{wraptable}

We ablate the skill bank to the self-teacher in OPSD.
Table~\ref{tab:opsd-bank-controls} changes only the privileged context with other settings fixed. An empty privileged context
leaves it at the outcome-only level, so the distillation mechanism on its own does not produce the gain. Scrambling the words of
the \xtree{} bank, which keeps its vocabulary but destroys its chains, drops $0.6$.  The \xtree{} remains the most effective way to maintain the useful information from a trajectory pool with no LLM call, the comparison
Figure~\ref{fig:webshop-opsd} draws against the LLM-written bank.

\section{Stability and scope: the spread behind each mean}
\label{app:stability}

The main tables carry means only for better illustration. This section states how each mean is formed and how stable it is. All results are means over three seeds of the same configuration, same as compared methods and prior work.
A paired comparison takes the per-run difference between a run and its control and reports its mean $\Delta$ together with $t = \Delta / (\mathrm{sd}(\Delta)/\sqrt{n})$ over the $n$ paired runs.

\textbf{ScienceWorld.} Table~\ref{tab:stability} gives the mean and standard deviation (std) of the runs that enter each cell. The std ranges
\begin{table}[t]
\centering
\caption{\textbf{Spread behind ScienceWorld.} SR (\%) as
mean $\pm$ sample standard deviation behind Figure~\ref{fig:online-rlvr} and the OPSD results of Figure~\ref{fig:webshop-opsd} (top) and
Table~\ref{tab:rollout-n} (bottom, 1.5B).}
\label{tab:stability}
\footnotesize
\begin{tabular}{@{}llccc@{}}
\toprule
\textbf{Scale} & \textbf{Setting} & \textbf{G0} & \textbf{G1} & \textbf{held-out G2} \\
\midrule
1.5B & Outcome-only GRPO & $38.6 \pm 2.5$ & $36.2 \pm 0.3$ & $22.8 \pm 1.3$ \\
 & \quad + \xtree{} bonus & $39.8 \pm 1.0$ & $39.5 \pm 1.7$ & $25.6 \pm 1.3$ \\
 & \quad + OPSD, LLM-written bank & $39.4 \pm 1.7$ & $38.4 \pm 2.1$ & $25.4 \pm 2.0$ \\
 & \quad + OPSD, \xtree{} bank & $38.7 \pm 2.7$ & $35.9 \pm 1.8$ & $26.3 \pm 1.8$ \\
3B & Outcome-only GRPO & $54.6 \pm 6.0$ & $48.8 \pm 3.1$ & $24.0 \pm 3.8$ \\
 & \quad + \xtree{} bonus & $56.4 \pm 3.7$ & $52.6 \pm 5.6$ & $26.2 \pm 2.3$ \\
 & \quad + OPSD, LLM-written bank & $54.3 \pm 1.4$ & $48.6 \pm 2.2$ & $21.6 \pm 0.9$ \\
 & \quad + OPSD, \xtree{} bank & $58.0 \pm 7.7$ & $51.8 \pm 5.4$ & $24.1 \pm 1.5$ \\
7B & Outcome-only GRPO & $56.7 \pm 8.4$ & $50.2 \pm 3.6$ & $21.7 \pm 1.6$ \\
 & \quad + \xtree{} bonus & $61.6 \pm 8.7$ & $53.1 \pm 4.6$ & $25.6 \pm 1.5$ \\
 & \quad + OPSD, LLM-written bank & $60.4 \pm 6.5$ & $52.8 \pm 5.1$ & $22.3 \pm 2.8$ \\
 & \quad + OPSD, \xtree{} bank & $62.5 \pm 5.1$ & $54.7 \pm 4.5$ & $24.7 \pm 1.9$ \\
\midrule
\textbf{Budget} & \textbf{Setting} & \textbf{G0} & \textbf{G1} & \textbf{held-out G2} \\
\midrule
$n{=}2$ & Outcome-only GRPO & $12.9 \pm 5.4$ & $13.7 \pm 6.8$ & $11.8 \pm 6.6$ \\
 & \quad + \xtree{} bonus & $14.6 \pm 3.2$ & $15.5 \pm 3.5$ & $13.0 \pm 2.0$ \\
$n{=}4$ & Outcome-only GRPO & $24.2 \pm 3.4$ & $23.9 \pm 3.3$ & $12.1 \pm 7.8$ \\
 & \quad + \xtree{} bonus & $28.0 \pm 0.9$ & $27.5 \pm 1.3$ & $16.5 \pm 4.9$ \\
$n{=}16$ & Outcome-only GRPO & $47.5 \pm 2.6$ & $49.0 \pm 3.1$ & $26.6 \pm 3.2$ \\
 & \quad + \xtree{} bonus & $51.0 \pm 1.5$ & $47.9 \pm 2.2$ & $27.4 \pm 1.9$ \\
\bottomrule
\end{tabular}
\vspace{-10pt}
\end{table}

are widest at 7B and at the smallest rollout budgets based on outcome reward. The \xtree{}-based run is tighter in most runs, so the gain is stable, not bought with extra variance. The held-out G2 column is the noisiest at every scale.

\textbf{WebShop.} Table~\ref{tab:stability-webshop} gives every cell with
\begin{table}[t]
\centering
\caption{\textbf{Spread behind WebShop.} Success and graded score
(\%) as mean $\pm$ sample standard deviation behind Figure~\ref{fig:online-rlvr} and 
Figure~\ref{fig:webshop-opsd}. $\Delta$ ($t$) is
the paired success difference against the outcome-only row of the same scale
and its $t$-statistic.}
\label{tab:stability-webshop}
\small
\setlength{\tabcolsep}{5pt}
\begin{tabular}{@{}llccc@{}}
\toprule
\textbf{Scale} & \textbf{Setting} & \textbf{Success} & \textbf{Graded} & \textbf{$\Delta$ ($t$)} \\
\midrule
\multirow{4}{*}{1.5B}
 & RLVR, outcome-only & $73.8 \pm 1.9$ & $88.0 \pm 1.0$ & --- \\
 & \quad $+$ \xtree{} bonus & $76.5 \pm 1.6$ & $89.6 \pm 1.1$ & $+2.7$ (7.0) \\
 & OPSD, LLM-written bank & $74.7 \pm 1.9$ & $87.7 \pm 1.0$ & $+0.9$ (0.8) \\
 & OPSD, \xtree{} bank & $73.8 \pm 1.7$ & $87.4 \pm 1.1$ & $+0.0$ (0.1) \\
\midrule
\multirow{4}{*}{3B}
 & RLVR, outcome-only & $74.0 \pm 3.0$ & $87.7 \pm 1.6$ & --- \\
 & \quad $+$ \xtree{} bonus & $75.9 \pm 3.5$ & $88.4 \pm 1.6$ & $+1.9$ (2.4) \\
 & OPSD, LLM-written bank & $76.0 \pm 2.0$ & $88.2 \pm 1.2$ & $+2.1$ (2.6) \\
 & OPSD, \xtree{} bank & $77.1 \pm 1.6$ & $89.9 \pm 1.2$ & $+3.2$ (4.6) \\
\midrule
\multirow{4}{*}{7B}
 & RLVR, outcome-only & $73.8 \pm 3.9$ & $85.6 \pm 1.8$ & --- \\
 & \quad $+$ \xtree{} bonus & $77.4 \pm 1.9$ & $90.2 \pm 1.0$ & $+3.6$ (2.1) \\
 & OPSD, LLM-written bank & $78.5 \pm 3.0$ & $90.2 \pm 2.3$ & $+4.7$ (8.4) \\
 & OPSD, \xtree{} bank & $77.9 \pm 2.7$ & $90.0 \pm 1.5$ & $+4.1$ (3.1) \\
\bottomrule
\end{tabular}
\vspace{-10pt}
\end{table}

its spread. Std ranges from $1.6$ to $3.9$ points on success and $1.0$
to $2.3$ on the graded score. The \xtree{} bonus is ahead of 
outcome-only on every evaluation run of every cell of Figure~\ref{fig:online-rlvr}, which is the property a mean alone does not show. \begin{table}[t]
\centering
\footnotesize
\setlength{\tabcolsep}{5pt}
\caption{{Resource-matched controls and mining-corpus size, with their spreads} (WebShop, 1.5B).
Success (\%) as mean $\pm$ sd over three runs. \emph{Top}: both runs of a row share the warm start and the RL data, and \xtree{} is mined from the SFT trajectory.
``Anchor'' is the warm start's own success before RL. \emph{Bottom}: the \xtree{} is mined from $N$ random trajectories.}
\label{tab:webshop-controls-full}
\label{tab:corpus-size-full}
\begin{tabular}{@{}lccc@{}}
\toprule
 & \textbf{Anchor} & \textbf{outcome-only} & \textbf{$+$\xtree{}} \\
\midrule
\multicolumn{4}{@{}l}{\emph{\xtree{} mined from the warm start trajectories}} \\
500 samples & 24.2 & $73.8 \pm 1.9$ & $74.9 \pm 1.0$ \\
1{,}016 successful & 33.1 & $74.9 \pm 1.5$ & $78.1 \pm 2.1$ \\
1{,}824 all & 32.5 & $76.4 \pm 4.1$ & $77.4 \pm 1.1$ \\
\midrule
\multicolumn{4}{@{}l}{\emph{$N$ trajectories in mining corpus. All trained from the $500$-sample warm start}} \\
$N{=}50$, 9 skills & - & - & $68.7 \pm 0.7$ \\
$N{=}100$, 11 skills &- &- & $75.2 \pm 1.9$ \\
$N{=}200$, 16 skills &- & -& $72.9 \pm 2.1$ \\
$N{=}500$, 24 skills &- &- & $74.4 \pm 0.8$ \\
$N{=}1{,}824$ (full), 48 skills & -&- & $\mathbf{76.5} \pm 1.6$ \\
\bottomrule
\end{tabular}
\end{table}

Table~\ref{tab:webshop-controls-full} adds the spreads behind
Tables~\ref{tab:webshop-controls} and~\ref{tab:webshop-variants}, showing the stability of our results. 

\textbf{WebArena.} For the main results (Table~\ref{tab:webarena-main}), the spread is $0.44$ to $1.15$ points of SR. Our full recipe is $22.9 \pm 0.6$ (passes $23.5/22.8/22.3$) against $18.4 \pm 0.6$ for the SFT-Full reference
($18.7/18.7/17.7$), so the $+4.5$ gap is about seven times the larger of the two spreads.

\end{document}